\documentclass{article}

\usepackage{times}
\usepackage{graphicx}
\usepackage{amsmath}
\usepackage{amssymb}

\begin{document}
\title{How \textit{intelligent} are convolutional neural networks?}
\author{Zhennan Yan\\
\and
Xiang Sean Zhou\\
Siemens Medical Solutions USA, Inc.\\
65 Valley Stream Parkway, Malvern, PA 19355, USA\\
{\tt\small xiang.zhou@siemens-healthineers.com}
}
\maketitle
\begin{abstract}
	Motivated by the Gestalt pattern theory in psychology and visual perception, and the Winograd Challenge for language understanding, we design synthetic experiments to investigate a deep learning algorithm's ability to infer simple (at least for human) semantic visual concepts, such as symmetry, counting, and uniformity, etc., from examples. 
	A visual concept is represented by randomly generated, positive as well as negative, example images. We then test the ability and speed of algorithms (and humans) to learn the concept from these images. The training and testing are performed progressively in multiple rounds, with each subsequent round deliberately designed to be more complex and confusing than the previous round(s), especially if the true meaning of the concept was not grasped by the learner. However, if the semantic concept was understood, all the deliberate tests would become trivially easy.
	Our experiments show that humans can often infer a semantic concept quickly after looking at only a very small number of examples (this is often referred to as an ``aha moment": a moment of sudden realization, inspiration, insight, recognition, or comprehension), and performs \textit{perfectly} during all testing rounds (except for careless mistakes), even on samples that lie completely outside of the training distribution. On the contrary, deep convolutional neural networks or DCNNs could \textit{approximate} some concepts statistically, but only after seeing many ($\times10^4$) more examples. And it will still make obvious mistakes, especially during deliberate testing rounds or on samples outside the training distributions. This signals a lack of true ``understanding", or in other words, a failure to reach the right ``formula" for the semantics. We did find that some concepts are easier for DCNN than others. For example, simple ``counting" (i.e., 1 to 3 objects) is more learnable than ``symmetry", while ``uniformity" (e.g., in terms of shape variation) or ``conformance" (e.g., in terms of group behavior) are much more difficult for DCNN to learn.
	To conclude, we propose an ``Aha Challenge" for visual perception, calling for focused and quantitative research on Gestalt-style machine intelligence using limited training examples.
\end{abstract}

\section{Introduction}

Recently, deep learning methods~\cite{lecun2015deep,schmidhuber2015deep} have improved state-of-the-art in many domains, such as speech recognition, visual object detection and recognition, machine translation, etc. Deep Convolutional Neural Network (DCNN) is one of the most successful deep learning architectures. It has been widely adopted by the research communities, since Krizhevsky et al. ~\cite{krizhevsky2012imagenet} used a DCNN to almost halve the error rate in the ImageNet competition in 2012. Since then, DCNN's have achieved great successes in various computer vision tasks, approaching or even surpassing human-level performance in some tasks \cite{taigman2014deepface,he2015delving}.

The advantages of DCNN include the hierarchy of self-learned features and its ability to learn complex non-linear functions via direct end-to-end training~\cite{lecun2015deep}. The hierarchy of features, representing low-level image details as well as high-level abstract properties, can be learned from the raw data without any hand-crafted manipulation. At the same time, the complex function, which bridges the gap between raw data and the learning task, is learned from the examples by the back-propagation algorithm. Usually, there are millions of weights in a DCNN model. Since the complex model could capture data variance very well, it could absorb a large amount of training samples to avoid over-fitting.
Comparing with conventional machine learning algorithms, deep learning has a clear advantage in terms of easy implementation, seemingly unlimited learning capacity, and unprecedented performance, which makes it very attractive to both research community and industry in this era of big data. However, due to the end-to-end training of the ``black-box" layers, the general reasoning ability of DCNN is still not fully explored or understood, and the appropriate size of data set to train a DCNN is still mostly empirical.

The first CNNs only contain convolutional and pooling layers, which are directly inspired by the biological neural cells and the hierarchical architecture in animal visual systems \cite{hubel1962receptive,felleman1991distributed}. After years of evolution, CNNs have become deeper and deeper. More nonlinear and complex layers and structures are utilized in DCNNs, such as ReLU and normalization layers \cite{krizhevsky2012imagenet}, dropout layer \cite{srivastava2014dropout}, residual connection \cite{he2016deep}, inception modules \cite{szegedy2015googlenet,szegedy2016incep3} and so on. Complexity brought opacity---although there are ways to visualize what patterns might have been learned by intermediate layers \cite{zeiler2014visualizing,mahendran2015understanding}, thus shedding some lights on the inner working of a CNN, overall we still lack a thorough understanding of the learning process. Some studies have shown that carefully designed adversarial noises could mislead the learned model \cite{szegedy2013intriguing,evtimov2017robust}, casting doubt on the generalization capability of these new models---
Did they really \textit{learn}? \textit{what} did they learn?

 
In this paper, we use simple and well-defined concepts, such as ``symmetry", ``counting", and ``uniformity", and synthetic and clean examples to test and compare the ``intelligence" of algorithms and humans. The relationship between the number of training data and the classification accuracy is used to evaluate the learning speed. Unseen test data, especially those drawn from outside the training distribution, are used to evaluate the degree of learning at the semantic level. Implicit ``end-to-end" learning of such concepts is a stepping stone to the \textit{scaling up} of artificial intelligence for many real-world tasks such as diagnostic imaging, where symmetry (e.g., of the brain), counting (e.g., of vertebrae), and uniformity (of tissue texture or anatomical structures) are key features for the detection of certain diseases. 


\section{Related work}



There have been some attempts to analyze the complexity and learning capacity of artificial neural networks over the last decades \cite{giles1987learning,bishop1995neural}. Recently, Basu et. al. \cite{basu2016theoretical} derived upper bounds on the VC dimension of CNN for texture classification tasks. Szegedy et. al. \cite{szegedy2013intriguing} reported counter-intuitive properties of neural networks, and found adversarial examples with hardly perceptible perturbation that could mislead the algorithm. Since then, many more successful attacks at deep learning were reported \cite{papernot2016practical,behzadan2017vulnerability,huang2017adversarial}.

Deep networks' vulnerability to adversarial examples led to active research for a defense mechanism. Goodfellow et. al. \cite{goodfellow2014explaining} proposed adversarial training, and Hinton et. al. \cite{hinton2015distilling} proposed to distill the knowledge by model compression. Goodfellow et. al. \cite{NIPS_GAN} also proposed a zero-sum game framework for estimating generative models via an adversarial process, namely Generative Adversarial Nets (GAN). A GAN balances an adversarial data generator and a discriminator during training. And the trained discriminator is more tolerate to adversarial examples as a result.

While GAN related research reveals and repairs a ``statistical" weakness of CNNs, our study focuses on ``semantic" level limitation of CNNs. In other words, we ask ``how well can a CNN discover patterns or concepts from data" --- a capability that is a hallmark of natural intelligence. For example, can a CNN learn the concept of ``Symmetry" (we will start with the simplest ``bilateral symmetry") from examples? how fast (i.e., with how many training samples) can it learn it? and how general does it understand the concept? and ultimately, can the same network architecture be trained to learn another concept, e.g., ``counting"?

Few-shot learning \cite{snell2017prototypical,ravi2016optimization,lake2015human}, one-shot learning \cite{fei2006one} or even zero-shot learning \cite{palatucci2009zero,socher2013zero} are trying to adopt a classifier to accommodate new classes not seen in training, given only a few examples, one example, or no example at all, respectively. The goal is to transfer learned knowledge and make the model generalizable to new classes or tasks. 
However, the new patterns tested in these papers are mostly analogous or homologous to the learned patterns. They have not tested the kind of semantic Gestalt visual concepts, which are more diverse and more challenging for machines, but mostly trivially easy for humans. Nevertheless, zero-shot learning capability of DCNN was observed occasionally in some rounds within our experiments. For example, see the near 100\% performance on the ``Deliberate test 1" rows under Setting 2 and Setting 3 in Table~\ref{tab:count_object}.
There were work on heuristic program to solve visual analogy IQ test \cite{evans1964heuristic}, and explicit modeling of higher level visual concepts based on low level textons \cite{zhu2005textons,wu2007compositional}. We approach these topics from a different angle, using a classification problem to implicitly embed the concepts, and focus on the testing of the limit of end-to-end learning capacity of machines.

A similar question was posed in the language understanding domain by Winograd \cite{levesque2011winograd}, where a collection of questions can be easily understood, thus answered by a reasonably intelligent human, but not easily at all by an algorithm. The concepts selected in this study are also easily discoverable by a reasonably intelligent human, and our experiments confirm this quantitatively. One aspect that is unique to human visual perception process is that there is often an "Aha!" moment, after which the concept is fully understood, and error rate drops to zero or near zero immediately. This is not observed in algorithms. We believe the reason behind this cognitive gap is the same as that underlines the Winograd Schema Challenge, one that is related to the accumulative human experience (both environmental and societal) or some may call it ``general intelligence".

There have also been a long history of research in the classic computer vision domain called Gestalt visual perception theory \cite{wagemans2012century,jakel2016overview}. Our study draws upon some insights gained from this research field. However, with a few simple concepts, we are only scratching the surface. As future work, there are many interesting concepts to explore, such as similarity or uniformity, continuity or conformance, and proximity or grouping, etc.

\section{Study design}


The inception v3 networks developed by Szegedy et al.~\cite{szegedy2016incep3} obtained high classification accuracy with relatively less computational cost. We use this network in this study as a representative method of DCNNs. 

Three types of visual recognition/classification tasks are studied in the following sections. The first one is based on the concept of ``symmetry" (section \ref{sec:symm}), the second one is based on ``counting" (section \ref{sec:count}), and the last one is based on ``grouping or conformance behavior" (section \ref{sec:group}). First two tasks include several sub-tasks, which may require additional learning of concepts such as ``uniformity" or ``grouping". All tasks are designed as binary classification problems.

We create synthetic data sets for these image recognition problems.
All images are generated in size of $200 \times 200$.


To handle the binary tasks and these synthetic images, we adapt the inception v3 model by changing the input size, and replacing the original softmax output layer by one hidden fully connected layer of 1024 nodes (with relu activation) plus one new softmax layer of 2 nodes. The weights of the network (except the new layers) are initialized from the pre-trained model on ImageNet database. 

\section{Symmetry}
\label{sec:symm}


In this section, we investigate the learnability of the concept ``symmetry" by CNNs as well as by humans. We focus on the simple form of bilateral symmetry, but test both global (in section \ref{sec:gloSym}) and local (in section \ref{sec:locSym}) symmetry. Global symmetry means that all the positive example images have bilateral symmetry, while local symmetry means that all positive images contains only symmetric shapes. Therefore, the local symmetry test is also a test of a ``uniformity" concept. 

To bring the study closer to real-world use cases, and to make it a bit more interesting, we conduct another test exploiting the symmetry in human faces (section \ref{sec:face}). 


\subsection{Global symmetry}
\label{sec:gloSym}

\begin{figure}
	\begin{center}
		\includegraphics[width=0.4\linewidth]{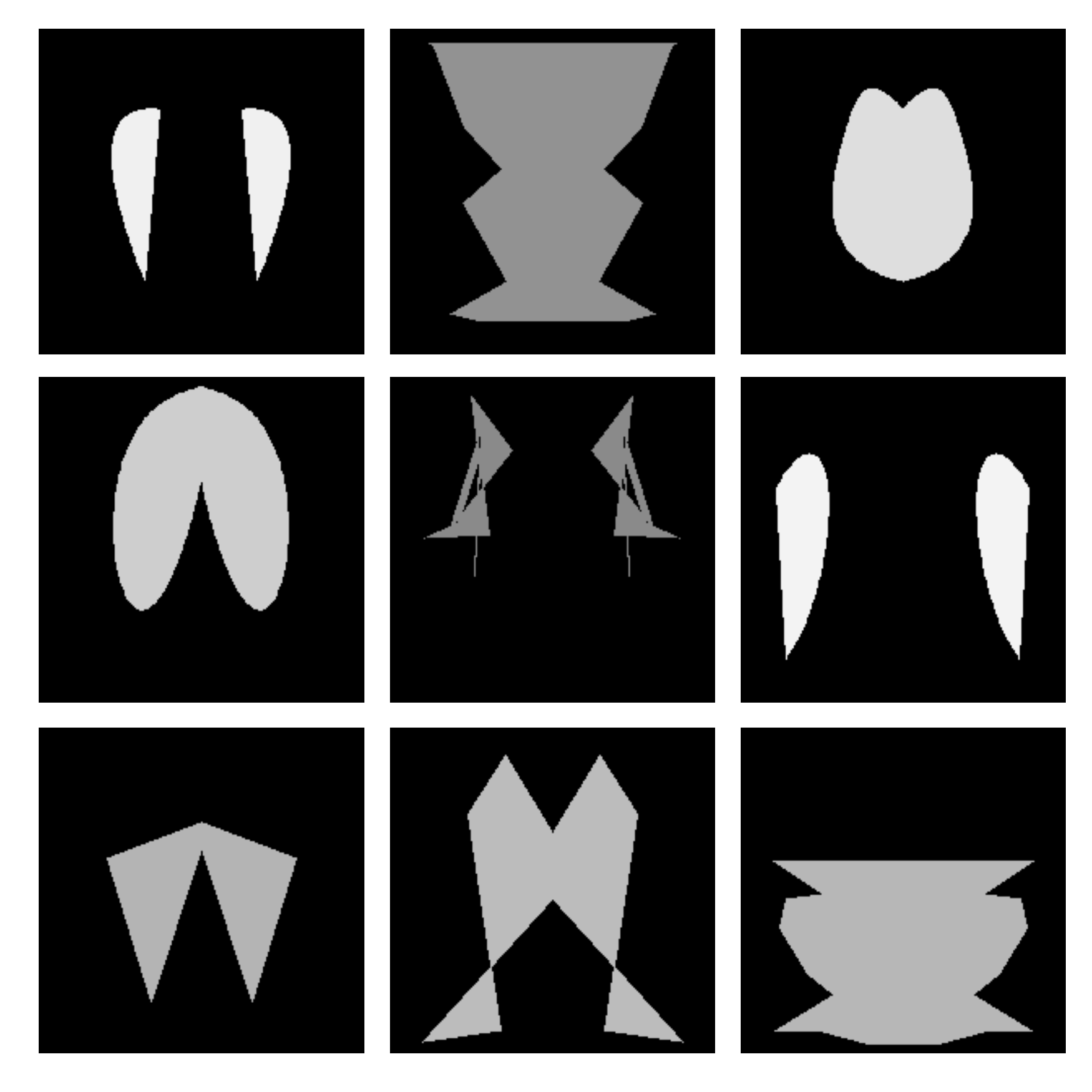}
		~~~\vspace{-5pt}
		\includegraphics[width=0.4\linewidth]{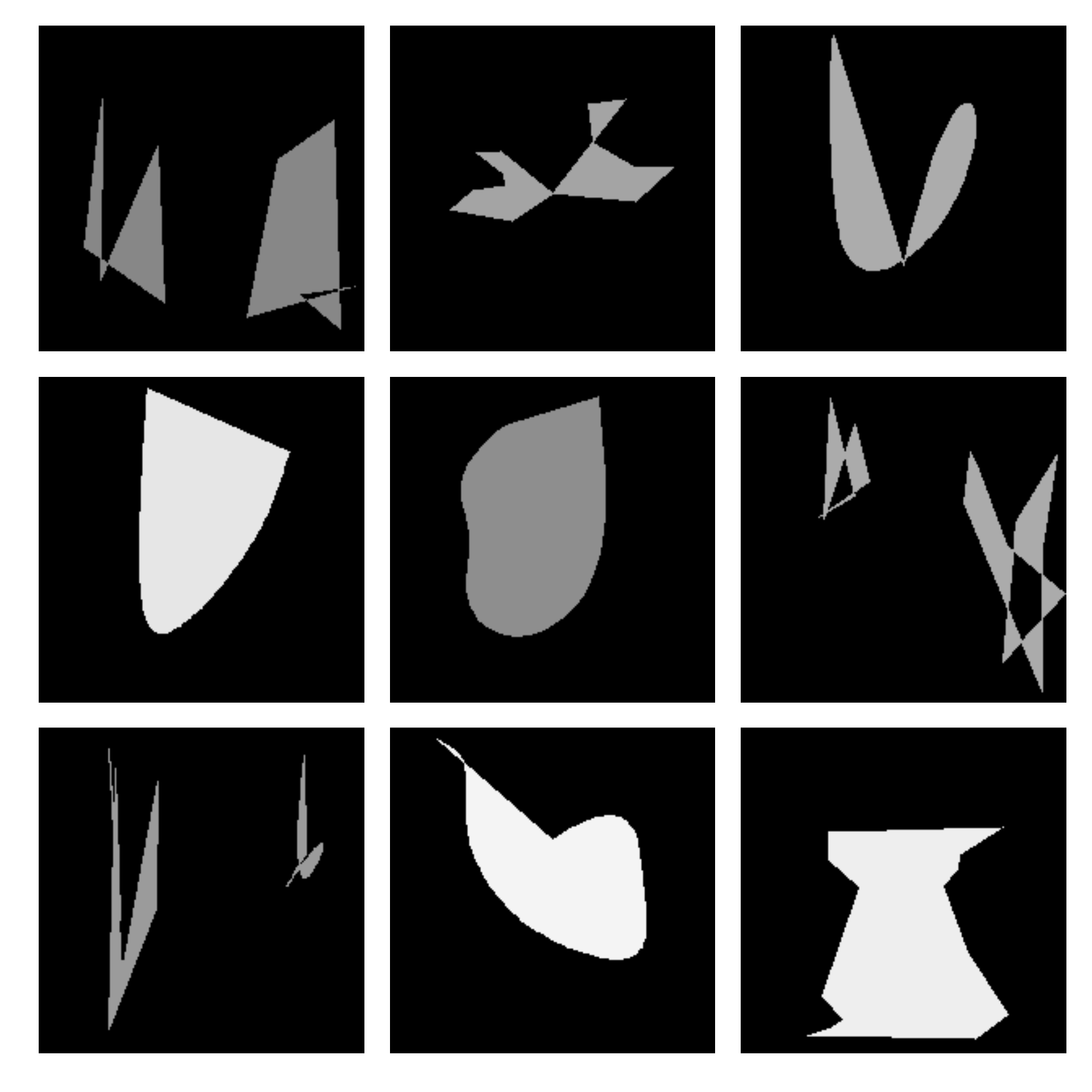}
	\end{center}
	\caption{Example of synthetic samples in $A_1$. Left: 9 symmetric samples; right: 9 asymmetric samples.}
	\label{fig:sample_symm1}
\end{figure}

Some examples of global symmetry are shown in Fig.~\ref{fig:sample_symm1}.
In this case, images are created by randomly sample control points and connecting them to form polygons. 
The interpolation between control points can be either a straight line or a bezier curve, which is determined randomly.
From these random shapes, symmetric images are generated by creating polygons from symmetrical control points or mirroring asymmetric shapes.

To program a computer algorithm \textit{explicitly} to detect such bilateral symmetry is trivial: just fold the image along the mid-line and then check the differences of overlapping pixels. However, for an algorithm to learn this seemingly simple operation \textit{implicitly}, from examples only, is not trivial at all --- especially if the benchmark is human performance. Humans can often grasp the concept quickly, after seeing only a few dozen examples.

We conduct four rounds of training and testing, designed to evaluate model generalization from different angles and at different levels. The first round is a statistical test within the same sample distribution as the training. This round is a ``smoke test" to ensure that the network is working. The second and third rounds of testing are deliberate and adversarial manipulation of the training set in order to test for ``understanding", or the lack thereof, of the real concept. The last round of test uses samples outside and far away from the training distribution, checking again for concept-level comprehension from another angle.

In the first round, we generate examples randomly similar to Fig.~\ref{fig:sample_symm1}. Denoting $A_1$ as the training set, $B_1$ as the validation set, and $C_1$ as the testing set, we can train model using $A_1$ and $B_1$, then test on $C_1$. 
To understand the generalization ability of the model \textit{on the concept level}, we also generate new ``deliberate" test samples $D_1$ based on $A_1$, with small but just enough modifications such that the class labels are all reversed. $D_1$ should be easy for a learner that has learned the concept, but very confusing for a learner that merely memorized (i.e., overfit to) $A_1$, because \textit{every sample in $D_1$ has a similar-looking counterpart in $A_1$}.

\begin{figure}
	\begin{center}
   \includegraphics[width=0.4\linewidth]{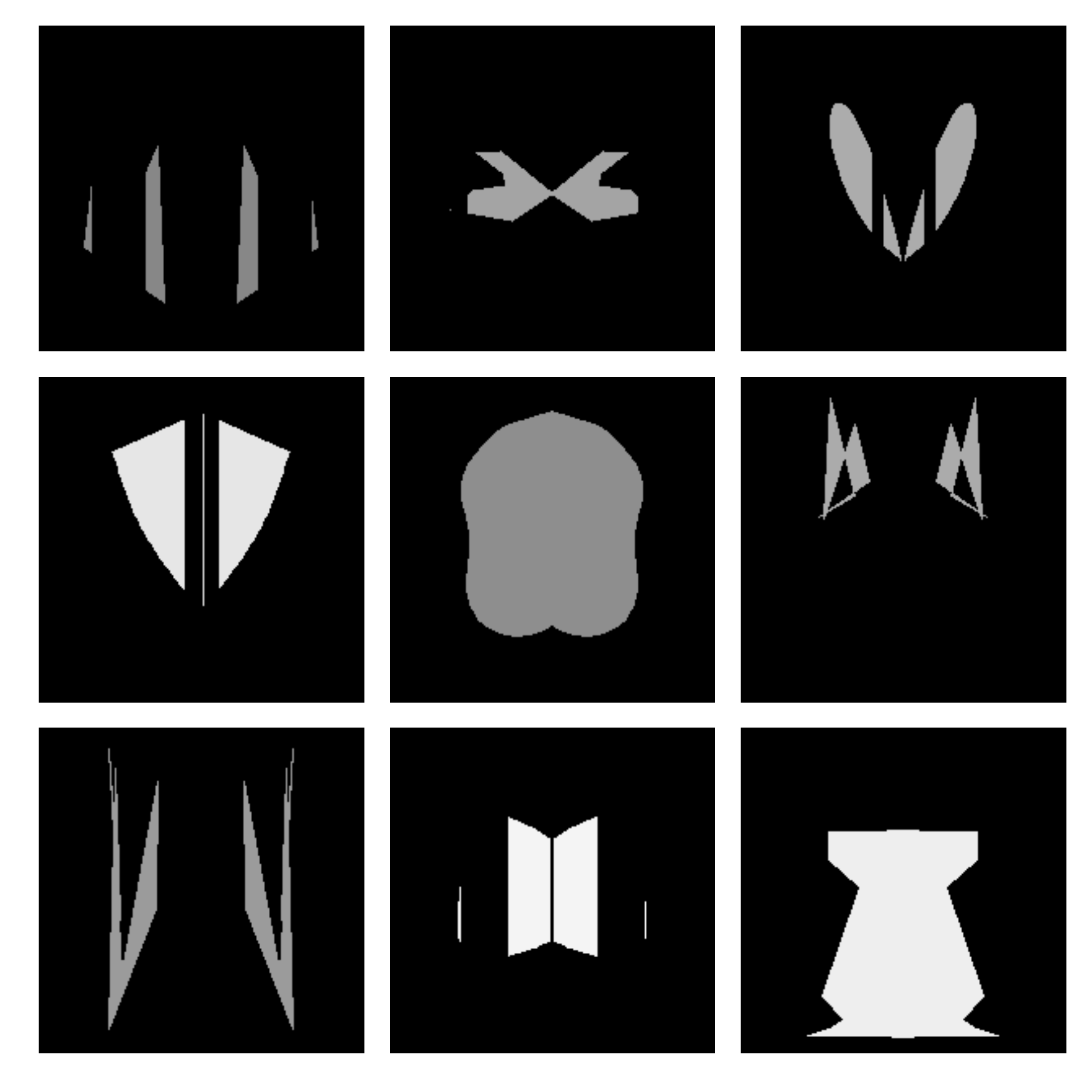}
   ~~~\vspace{-5pt}
	 \includegraphics[width=0.4\linewidth]{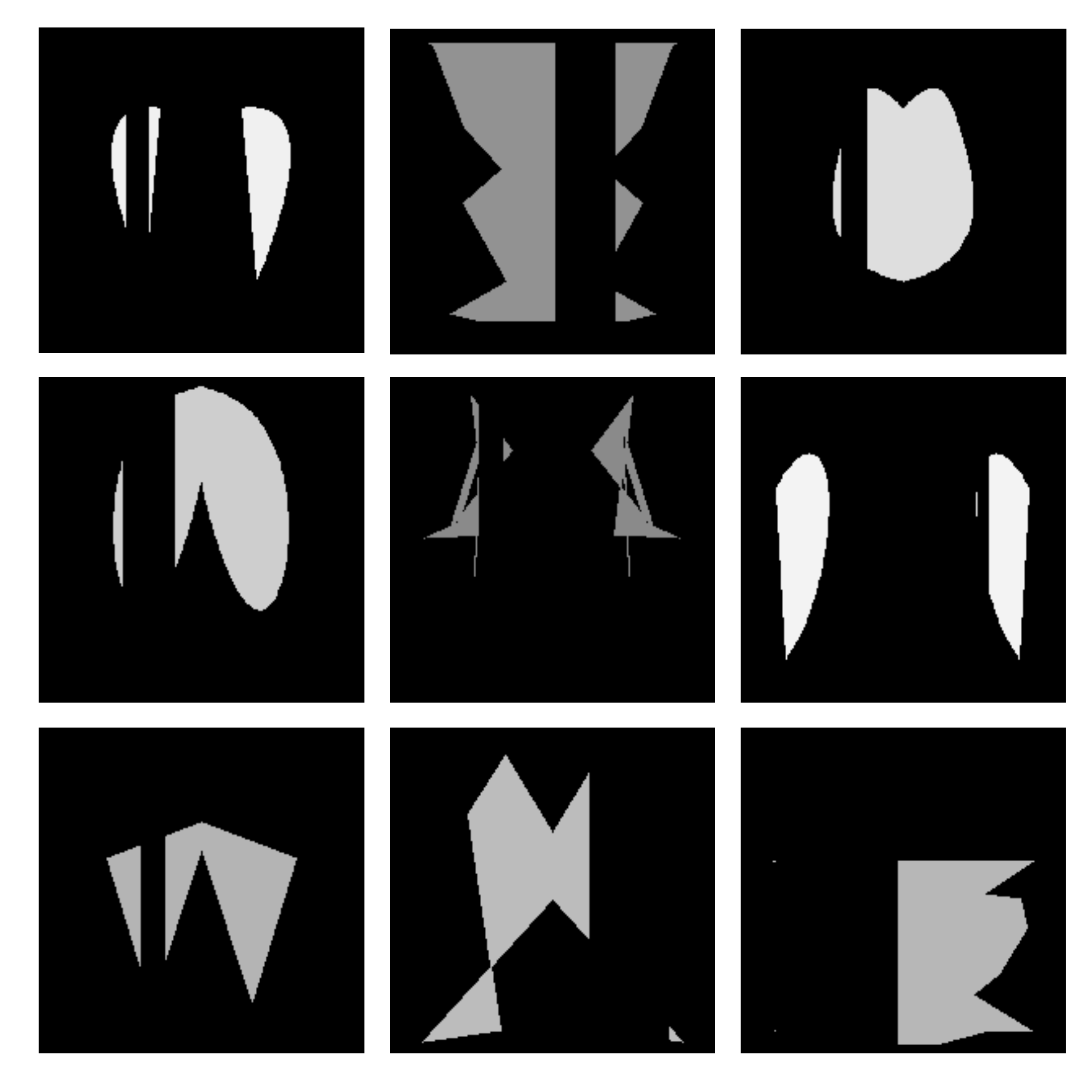}
	\end{center}
   \caption{Example of deliberate samples from $D_1(A_1)$. Left: 9 symmetric samples, which are generated from asymmetric ones in Fig.~\ref{fig:sample_symm1}; right: 9 asymmetric samples, which are generated from symmetric ones in Fig.~\ref{fig:sample_symm1}.}
\label{fig:sample_symm2}
\end{figure}

Denote $D(\cdot)$ as the operation to create deliberate test from existing samples. Various linear or non-linear functions can be applied in $D(\cdot)$. We use different operations in each round: $D_1$, $D_2$, and $D_3$. Specifically, $D_1(\cdot)$ on symmetric samples is removing some random part of the foreground in either left or right half to achieve asymmetry, while $D_1(\cdot)$ on asymmetric samples is mirroring one side of the image on the other side to achieve symmetry, and then, with a chance of $50\%$, symmetrically erasing some part of the image (this last step is to avoid bias by adding similar erasion pattern on both classes). Some of the samples in set $D_1(A_1)$ are shown in Fig.~\ref{fig:sample_symm2}.

\begin{figure}
\begin{center}
   \includegraphics[width=0.4\linewidth]{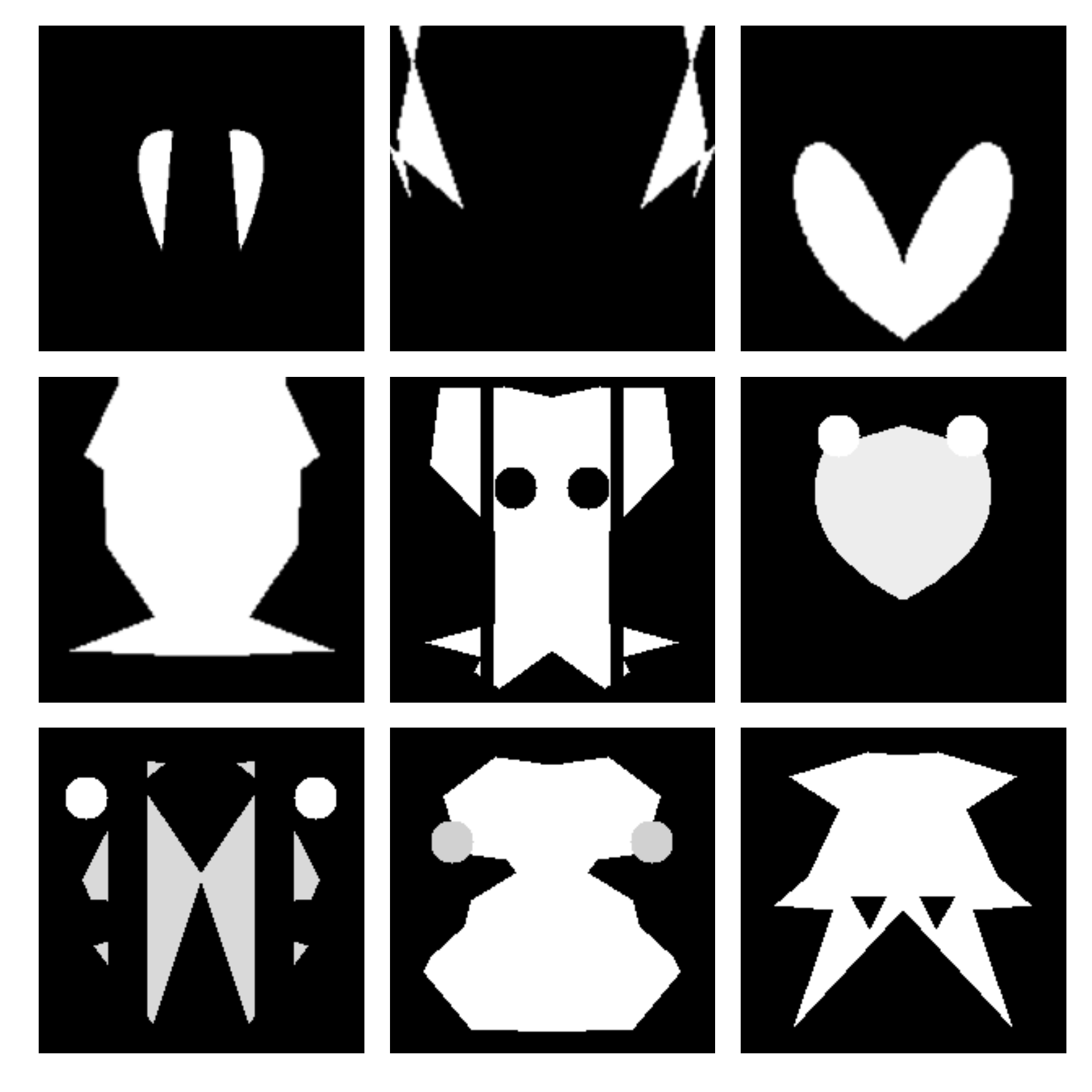}
   ~~~\vspace{-5pt}
	 \includegraphics[width=0.4\linewidth]{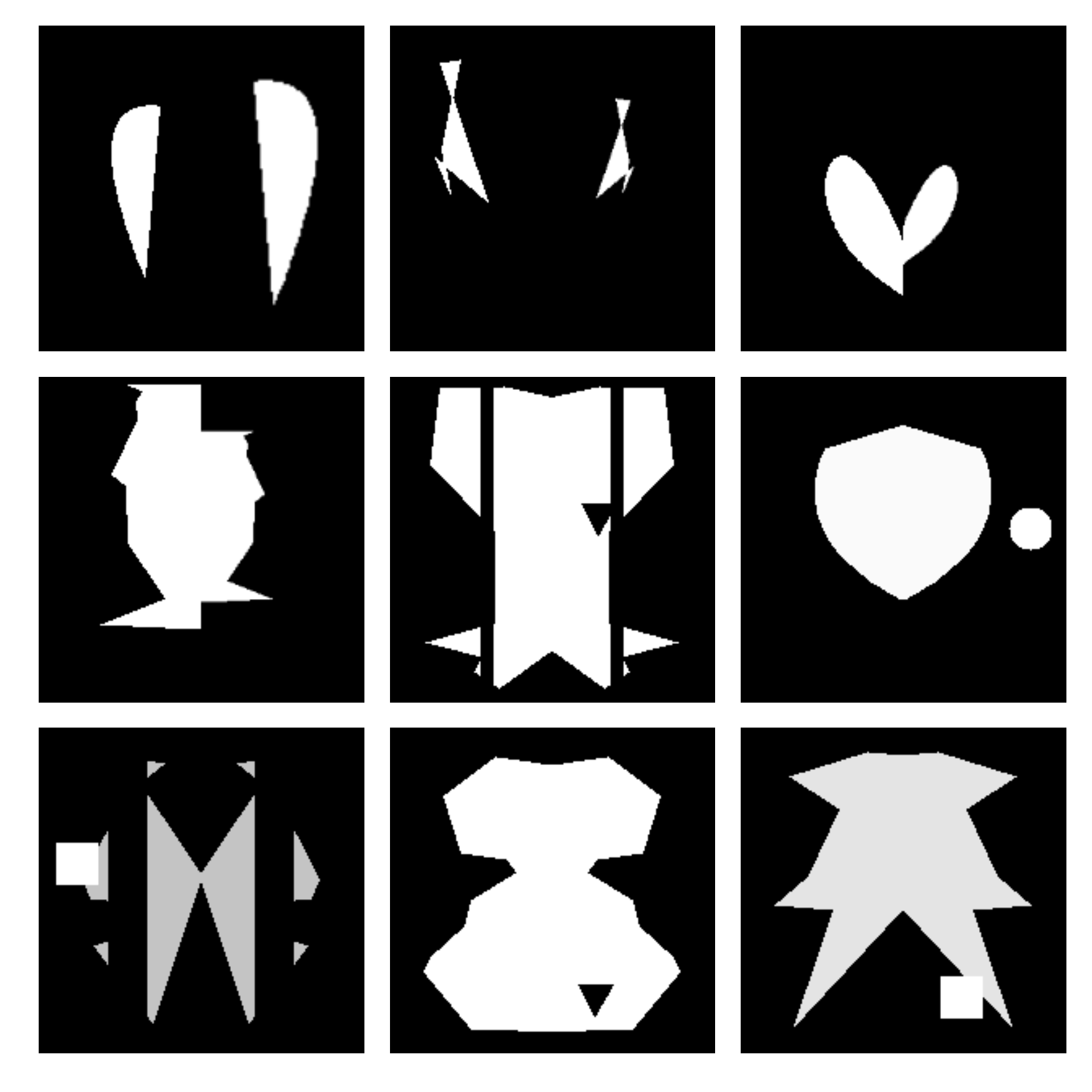}
\end{center}
   \caption{Example of 2nd round deliberate samples from $D_2(A_2)$. Left: 9 symmetric samples; right: 9 asymmetric samples.}
\label{fig:sample_symm3}
\end{figure}

In the second round, we construct a new training set $A_2={A_1\cup D_1(A_1)}$, validation set $B_2={B_1\cup D_1(B_1)}$, and train a new model using $A_2$ and $B_2$. Then, we test it on $D_2(A_2)$. The $D_2(\cdot)$ operation is based on scaling of, or adding small shape object(s) randomly (with a 50\% chance) to, the symmetric samples in $A_2$. More specifically, to create deliberate symmetric samples, we either scale the whole image, or add a pair of identical shapes at random but symmetric positions (in order to maintain symmetry). To create deliberate asymmetric samples, we either scale one side (left or right) of the image, or add a shape on one side. 
For scaling, we increase or decrease the size by $30\% \sim 50\%$. For the added shape, we evenly sample triangle, square, or ball, with the same size of $25$ pixels and random intensity $\in[128, 255)$. If the added small shape is located inside a foreground shape in the image, the intensity of the additional shape is set to 0 (making a hole). Random samples from $D_2(A_2)$ are shown in Fig.~\ref{fig:sample_symm3}.

\begin{figure}
\begin{center}
   \includegraphics[width=0.4\linewidth]{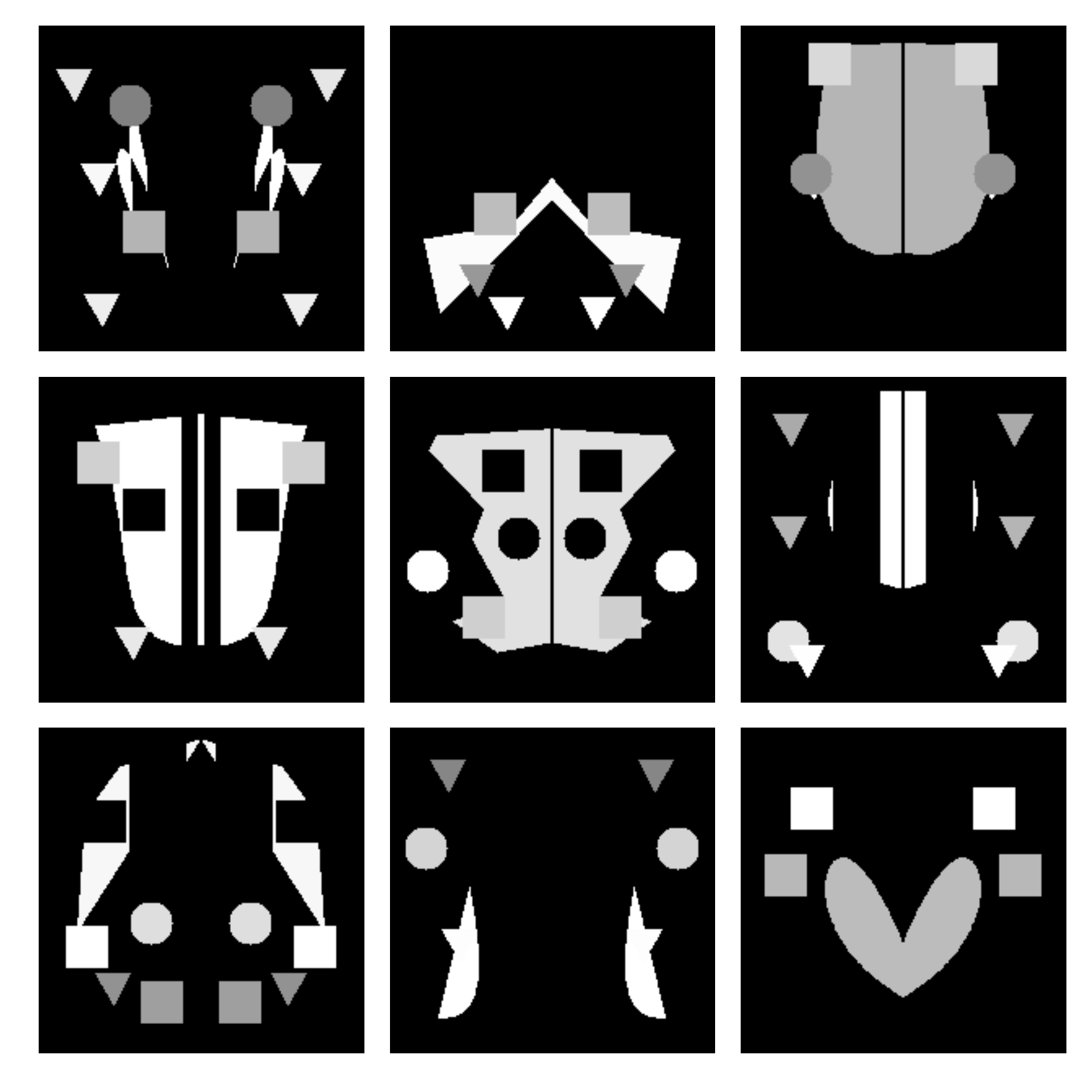}
   ~~~\vspace{-5pt}
	 \includegraphics[width=0.4\linewidth]{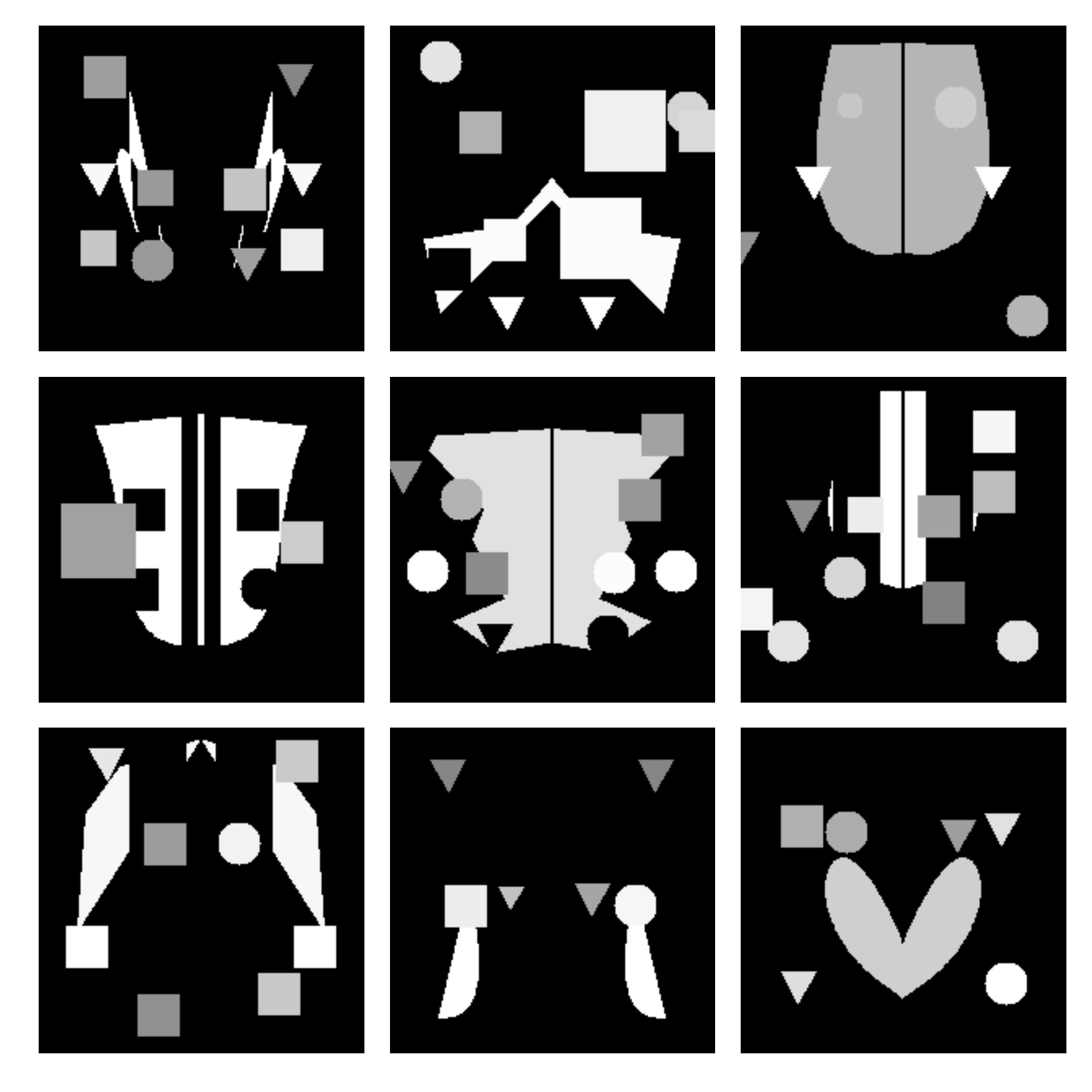}
\end{center}
   \caption{Example of 3rd round deliberate samples from $D_3(A_3)$. Left: 9 symmetric samples; right: 9 asymmetric samples.}
\label{fig:sample_symm4}
\end{figure}

In the third round, again we generate new training set $A_3={A_2\cup D_2(A_2)}$, and validation set $B_3={B_2\cup D_2(B_2)}$. New model is trained using $A_3$ and $B_3$, and tested on $D_3(A_3)$. Samples in $D_3(A_3)$ are created by adding more small shape objects into the symmetric images in $A_3$. The strategy of creating new symmetric samples is similar to $D_2(\cdot)$, but more objects/shapes are added. To create asymmetric samples, we randomly chose among three different approaches: (1) two random objects are added on the left and right sides, at asymmetric locations; (2) two random objects of different shape are added at symmetric positions; (3) two random objects of the same shape, but of different sizes, are added at symmetric positions. Examples from $D_3(A_3)$ are shown in Fig.~\ref{fig:sample_symm4}.

\begin{figure}
\begin{center}
   \includegraphics[width=0.4\linewidth]{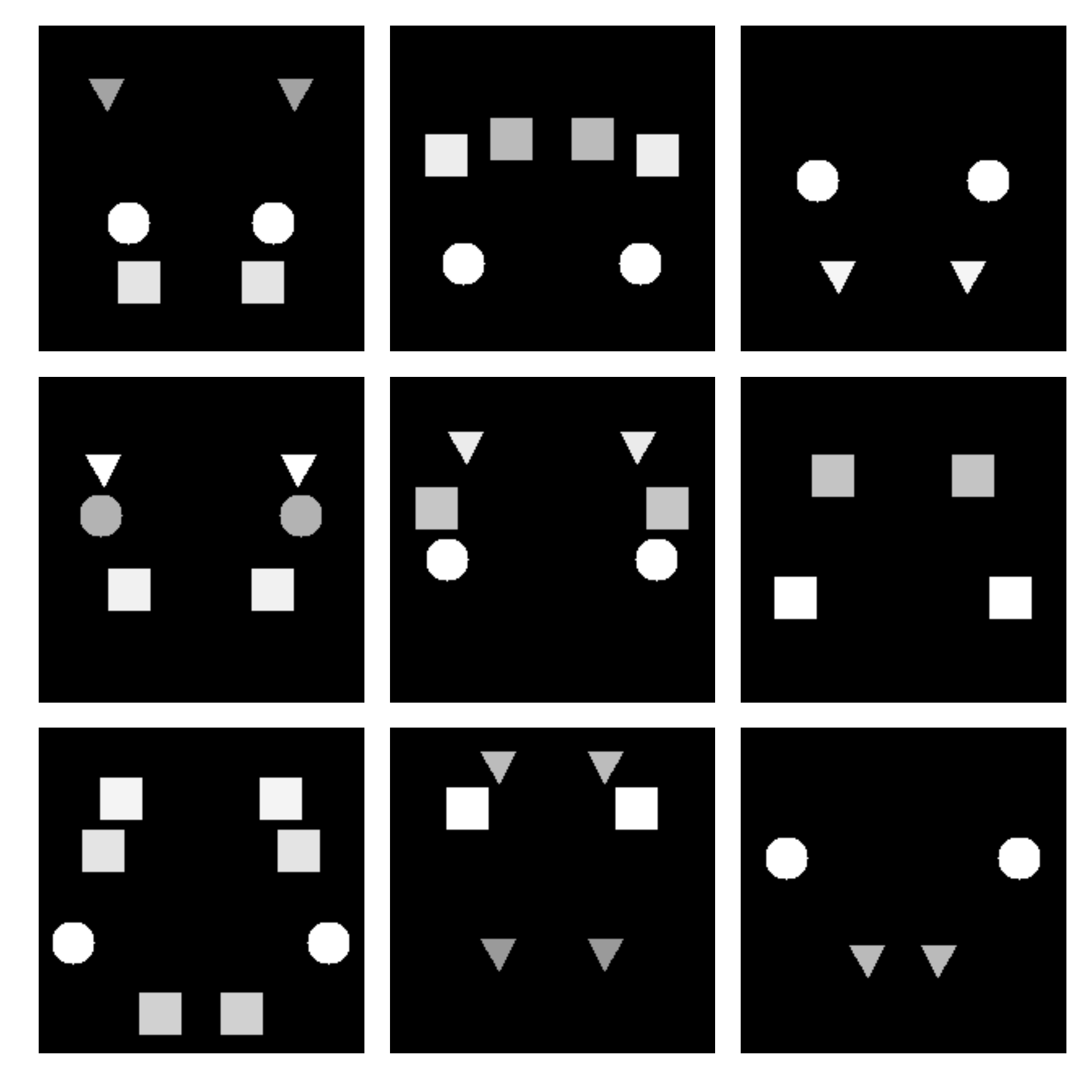}
   ~~~\vspace{-5pt}
	 \includegraphics[width=0.4\linewidth]{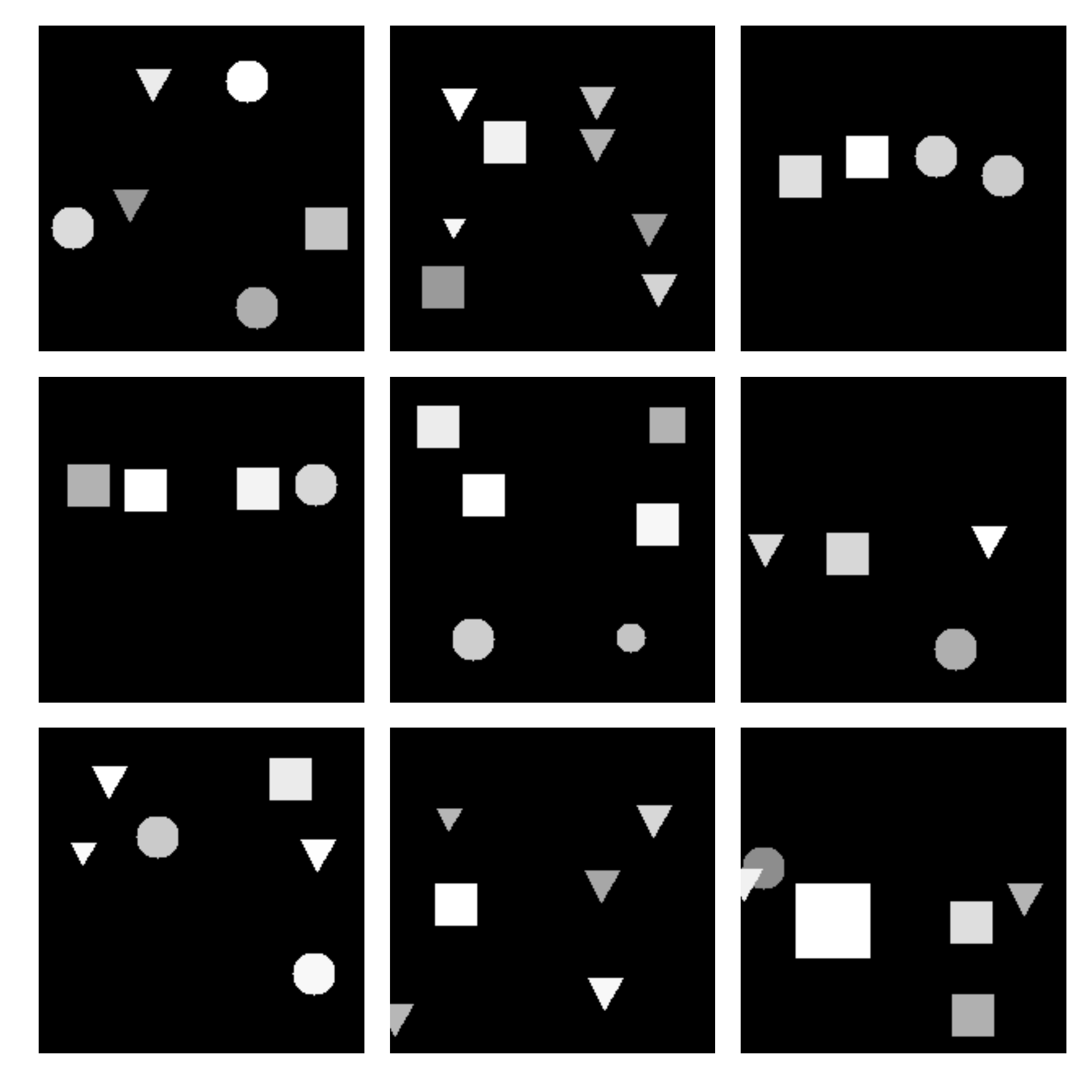}
\end{center}
   \caption{Example of 4th round samples from $A_4$. Left: 9 symmetric samples; right: 9 asymmetric samples.}
\label{fig:sample_symm5}
\end{figure}

In the last round, we generate brand new testing samples $A_4$ to evaluate the previous three models. These new samples are generated by placing shape objects (triangle, square and ball) at symmetric or asymmetric locations for the two classes. These samples lie completely outside of the previous training distributions, therefore, can serve as a good challenge for those models that have failed to learn the concept at the semantic level. Some samples from $A_4$ are shown in Fig.~\ref{fig:sample_symm5}. 

\subsection{Local Symmetry}
\label{sec:locSym}

\begin{figure}
\begin{center}
   \includegraphics[width=0.45\linewidth]{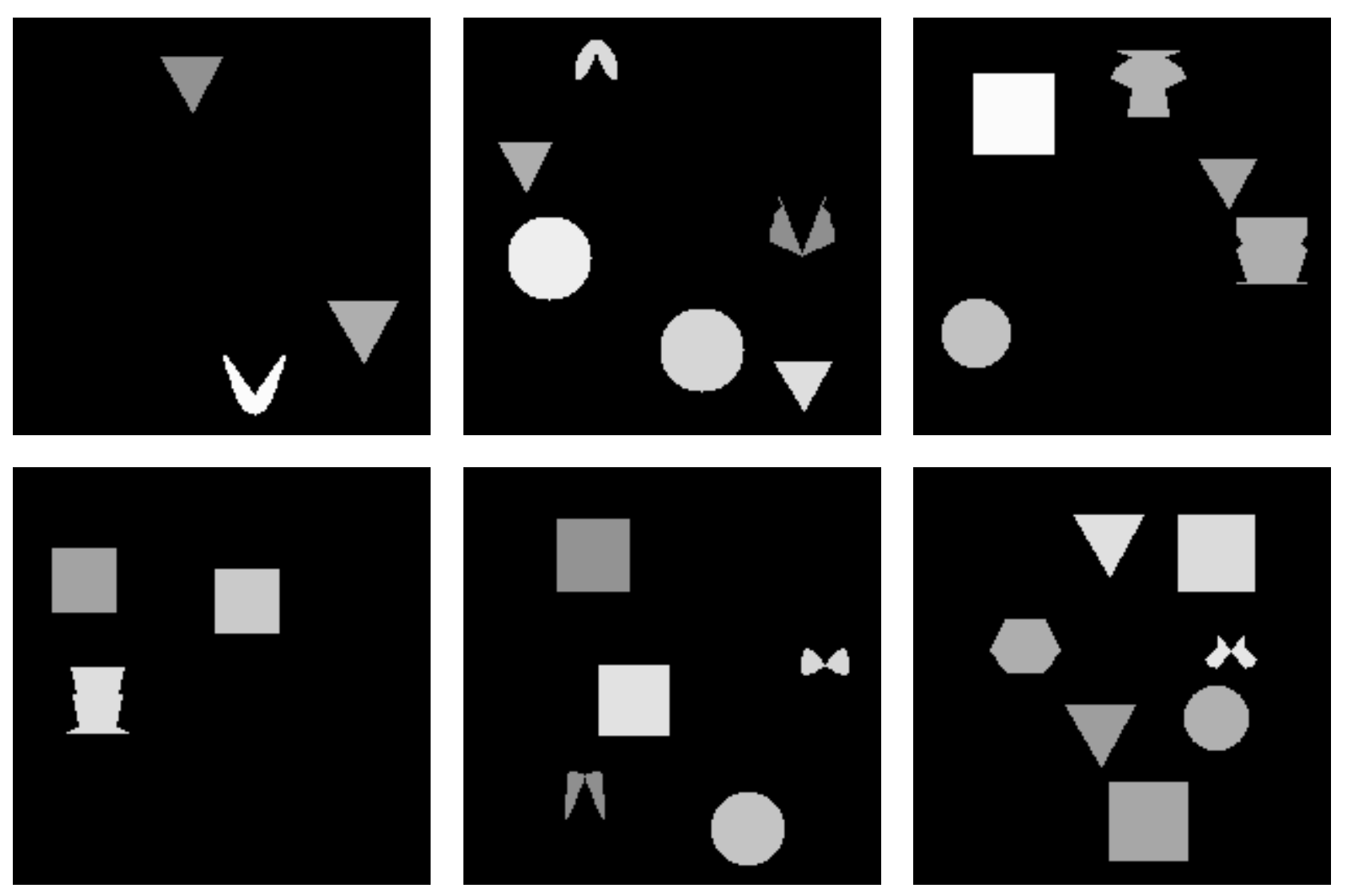}
   ~~~\vspace{-5pt}
	 \includegraphics[width=0.45\linewidth]{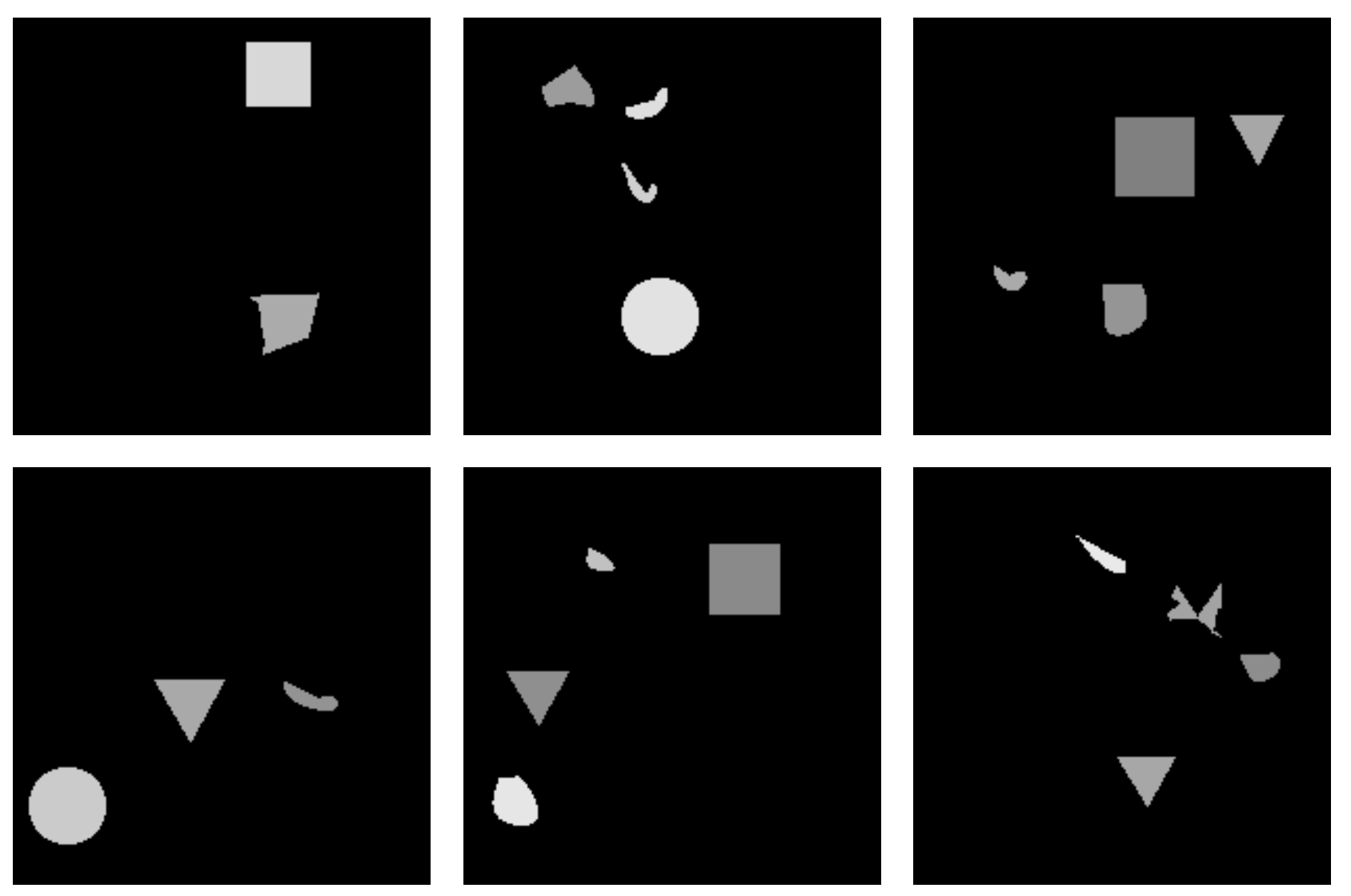}
\end{center}
   \caption{Example of synthetic training samples for local symmetry task. Left: 6 samples only containing symmetric patterns; right: 6 samples containing at least one asymmetric pattern.}
\label{fig:sample_symm_pat}
\end{figure}

\begin{figure}
\begin{center}
   \includegraphics[width=0.45\linewidth]{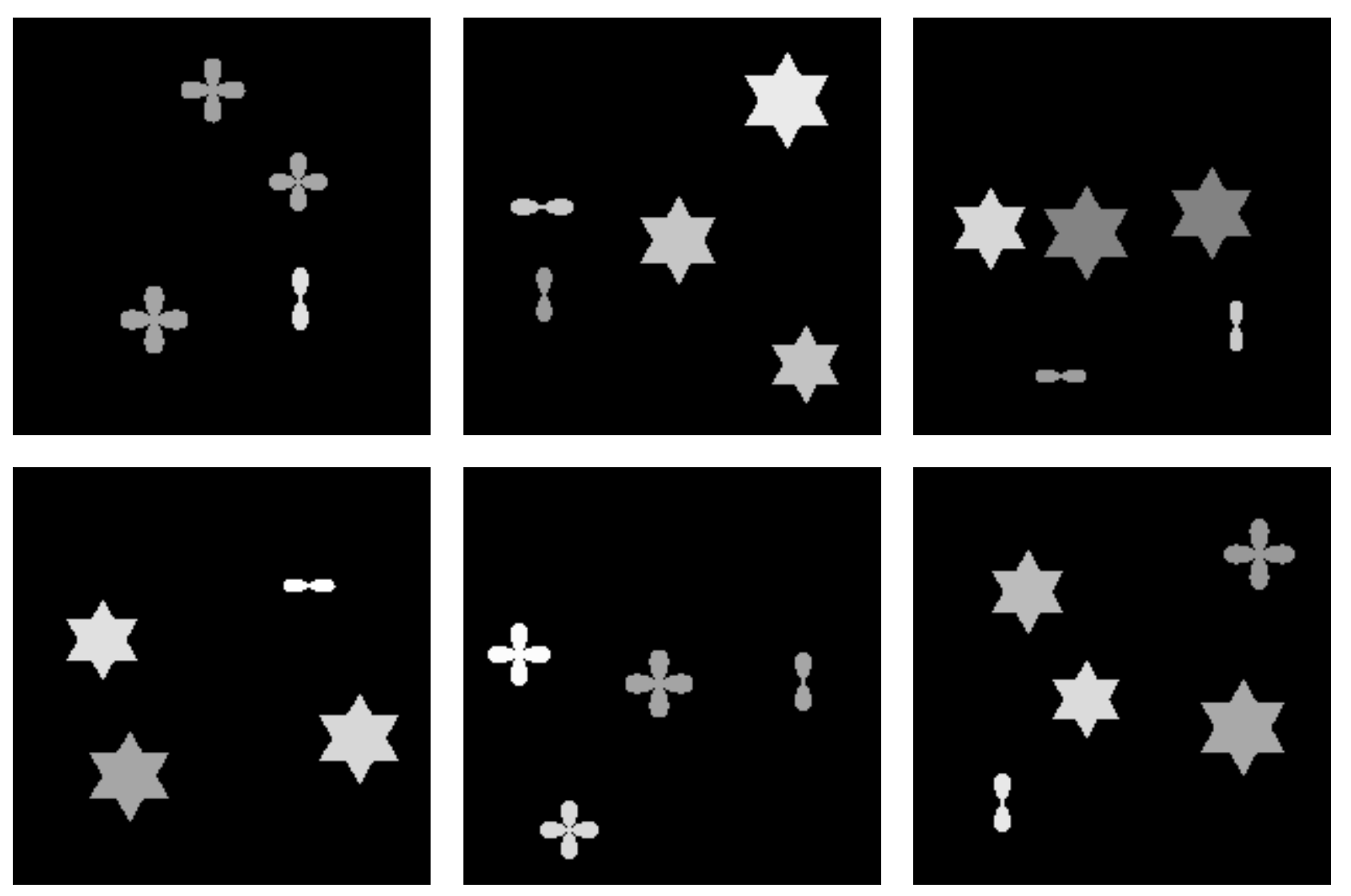}
   ~~~\vspace{-5pt}
	 \includegraphics[width=0.45\linewidth]{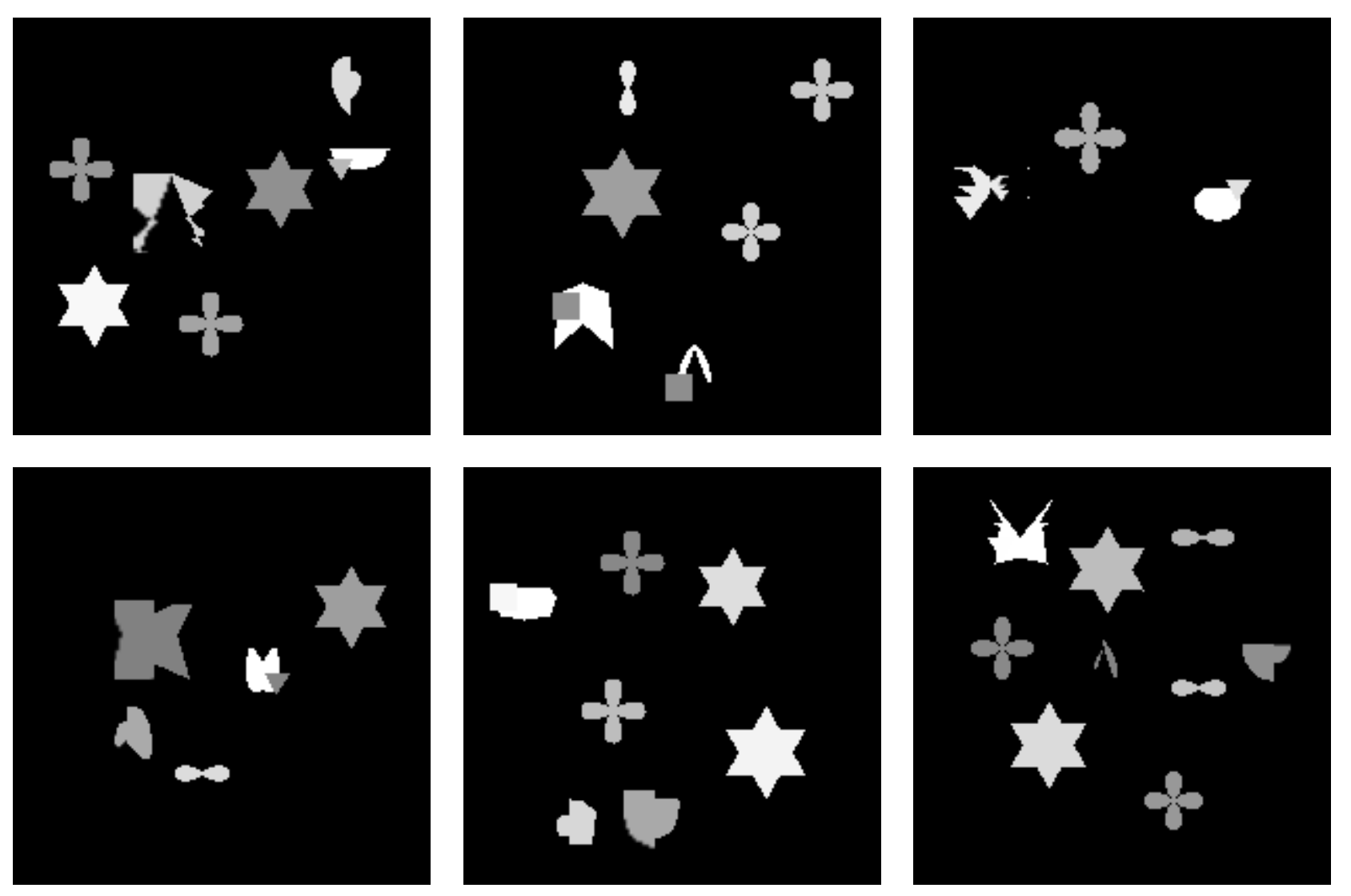}
\end{center}
   \caption{Example of deliberate testing samples for local symmetry task. Left: 6 samples only containing symmetric patterns; right: 6 samples containing at least one asymmetric pattern.}
\label{fig:sample_symm_pat2}
\end{figure}

In this sub-task, each image contains multiple objects. A positive image contains symmetric objects only, while a negative image contains at least one asymmetric object. Some examples are shown in Fig.~\ref{fig:sample_symm_pat}. 
In these image samples, objects include equilateral triangle, square, ball and connected-component polygon, which is generated in the similar way in $A_1$. Asymmetric objects are those asymmetric polygons. The sizes of objects are in range of $[30,40]$. No rotation is applied to objects to assure the bilateral symmetry at the object level.
After training, we create new deliberate samples to test the generalization. Two sets of deliberate samples are generated. The first testing set consists of new objects of the same size range. The symmetric objects include hexagram, 4-leaf flower (F4), and 2-leaf flower (F2). The asymmetric objects are created by using similar operation of $D_2(\cdot)$, which is scaling of, or adding object to, the symmetric polygon to make it asymmetric. Some examples are shown in Fig.~\ref{fig:sample_symm_pat2}.
The second testing set is constructed by using the training objects of larger sizes, in range of $[40,45]$.

\subsection{Normal/Tampered human face}
\label{sec:face}

The most obvious symmetry we see every day, and are very sensitive to, is probably in the human face, so we design an experiment to distinguish normal and tampered human faces. Although human faces are not completely symmetrical, higher degree of facial symmetry has been shown to be correlated with beauty, attractiveness~\cite{grammer1994human}, and personality \cite{fink2005facial}. Recently, deep learning has shown great power in face identification and verification \cite{taigman2014deepface,NIPS2014_5416}. In this sub-task, we aim to train and test DCNN's awareness to facial symmetry.

\begin{figure}
\begin{center}
   \includegraphics[width=\linewidth]{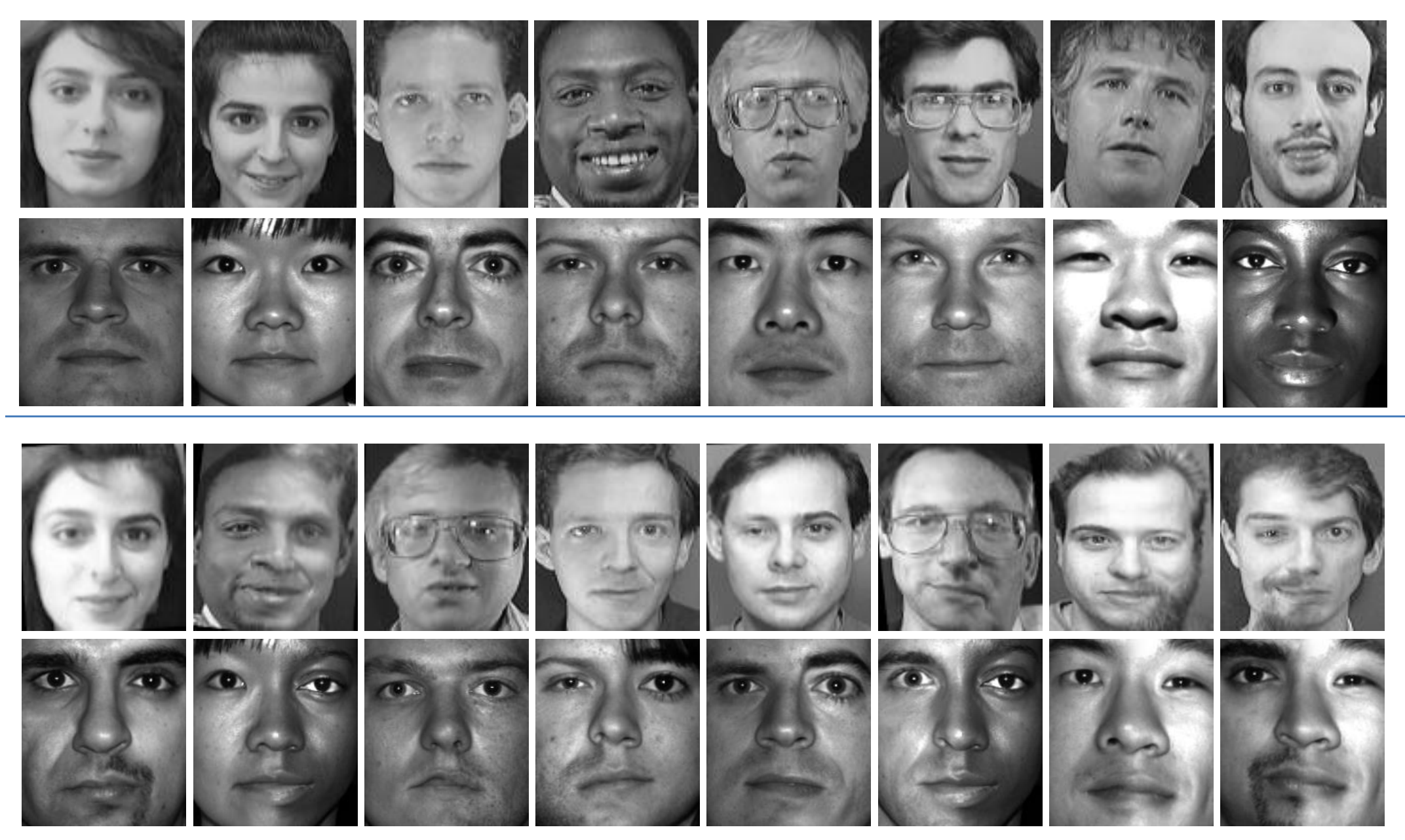}
\end{center}
   \caption{Example of real human faces (top 2 rows) vs. tempered faces (bottom 2 rows) in training set.}
\label{fig:sample_face1}
\end{figure}

We collected frontal face images from two public databases. 
Yale cropped face database\cite{GeBeKr01Yale,KCLee05Yale} and AT\&T face database\cite{samaria1994ATT}.
The tampered faces are created by fusing two half faces together (each half came from a different subject). Some examples are shown in Fig.~\ref{fig:sample_face1}.

To fuse two faces, we first use DLIB implementation of real-time face pose estimation algorithm \cite{kazemi2014one} to detect 68 facial landmarks. These landmarks are defined in \cite{sagonas2013300}. Secondly, we rotate each face so that the mid-line passing through the nose is vertical. Thirdly, we normalize the two faces to be fused to the same height, keeping the original width. Then, we align two faces together based on the face center (defined as mass center of eyes, nose and mouth). Finally, we merge two faces into one by using a sigmoid horizontal blending filter which is defined as:
\begin{equation}
\omega = \frac{1}{(1+exp(\frac{x-W/2}{\sigma}))},
\end{equation}
\begin{equation}
I(x,y) = \omega*I_1(x,y)+(1-\omega)*I_2(x,y)
\end{equation}
in which, (x,y) is the coordinates of image pixel; W is the image width; $I$, $I_1$ and $I_2$ are target image and two source images respectively; $\sigma=4$ in this study.

\section{Counting}
\label{sec:count}

The second major task in this study is counting. 
We design two sub-tasks related to counting in this section. The first one is simply counting objects. The second and also more difficult one is counting object types.

\begin{figure}
\begin{center}
   \includegraphics[width=0.46\linewidth]{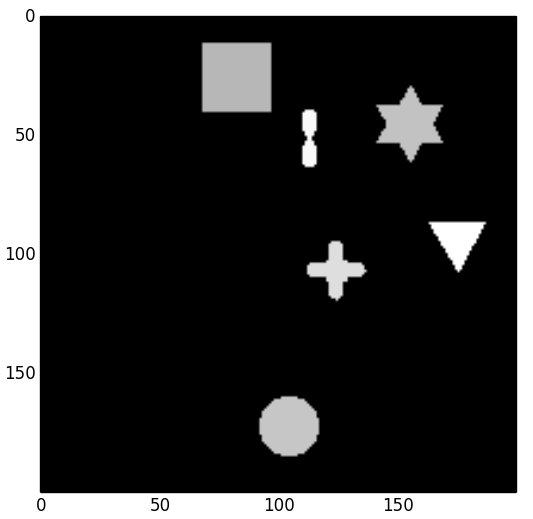}
\end{center}
   \caption{Example of synthetic objects.}
\label{fig:sample_shape}
\end{figure}

Six basic shapes are used to generate shape objects: equilateral triangle, square, ball, hexagram, 4-leaf flower (F4), and 2-leaf flower (F2). All shapes in an image can have different sizes, positions, intensities, and orientations in some cases. Fig. \ref{fig:sample_shape} shows a sample image. The size of each shape is measured by the longer edge of its bounding box.

\subsection{Counting objects}
\label{sec:count_obj}

To formulate counting problem as binary classification task, we design data set so that the two classes of image have different number of objects in each image sample.

\begin{figure}
\begin{center}
   \includegraphics[width=0.45\linewidth]{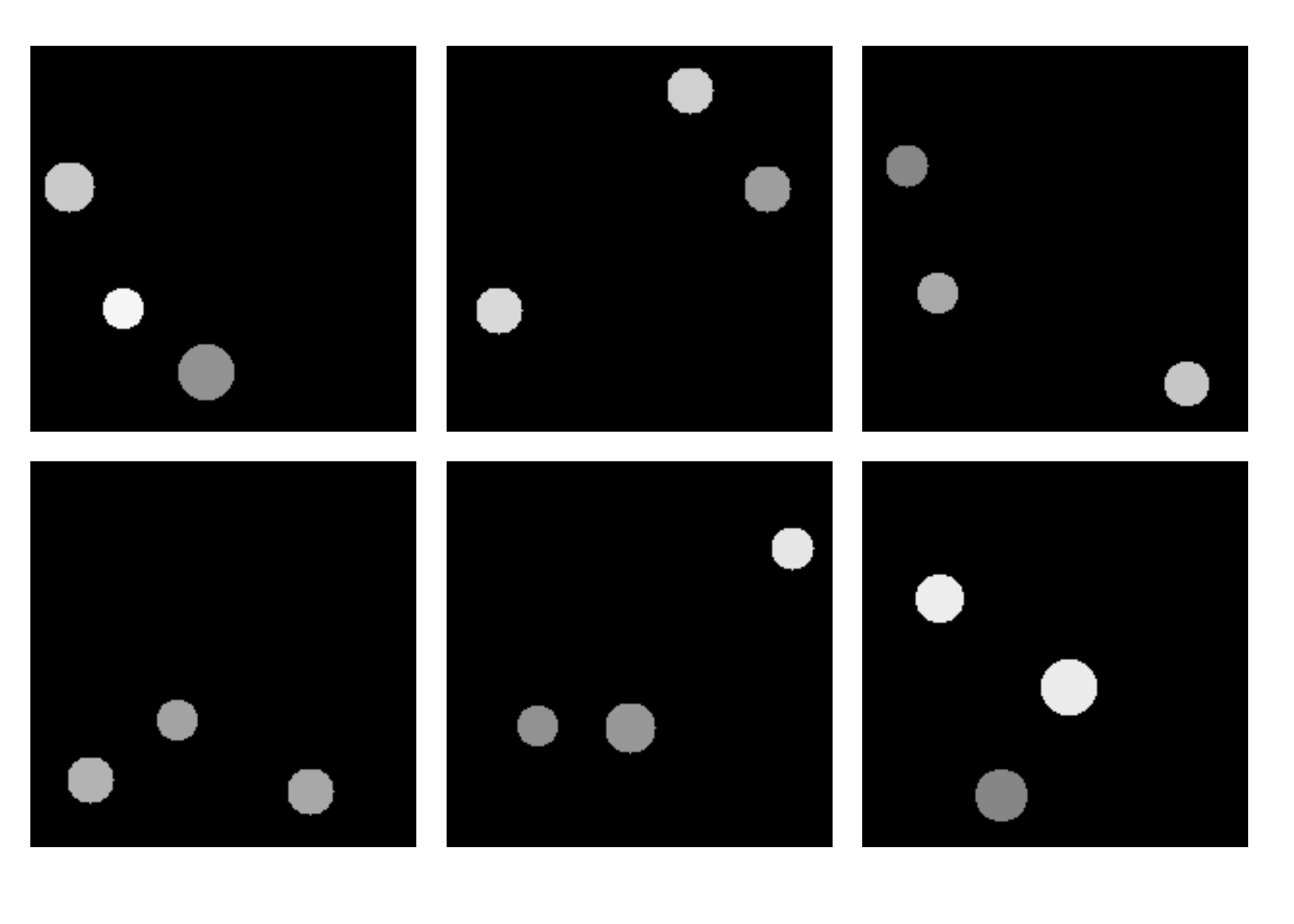}
   ~~\vspace{-10pt} 
	 \includegraphics[width=0.45\linewidth]{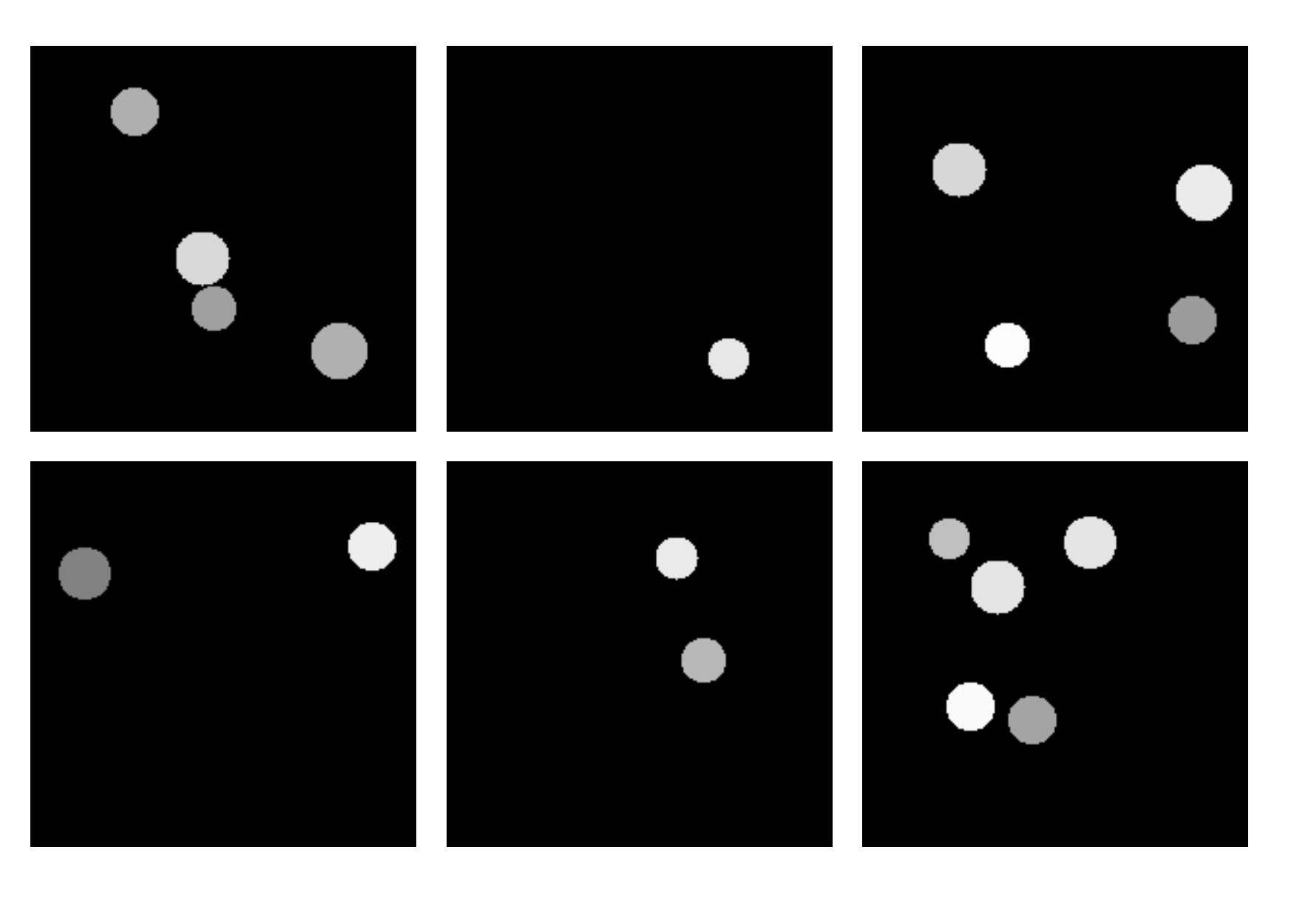}
	Setting 1\\ 
	 \includegraphics[width=0.45\linewidth]{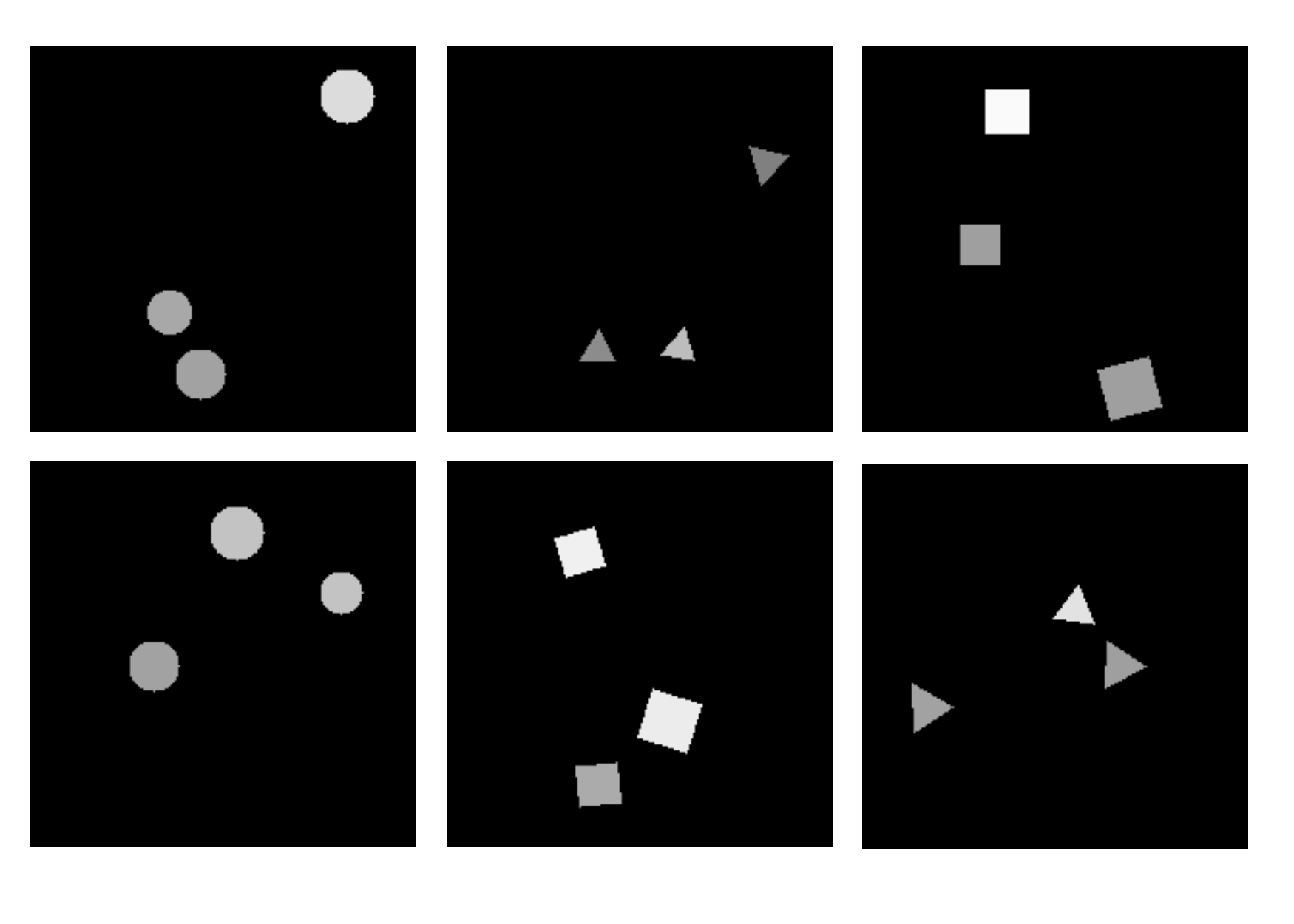}
   ~~\vspace{-10pt} 
	 \includegraphics[width=0.45\linewidth]{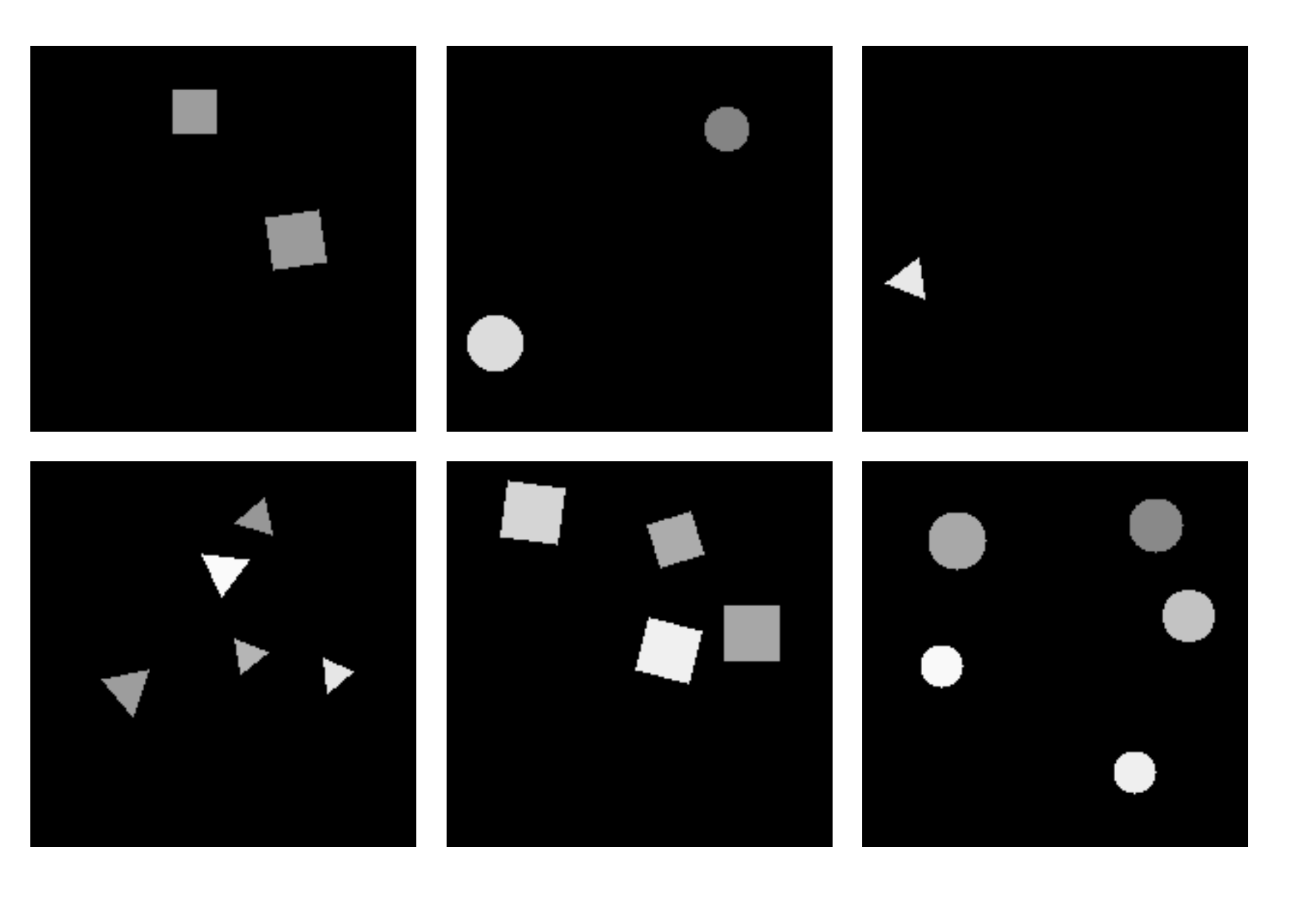}
	Setting 2\\
	 \includegraphics[width=0.45\linewidth]{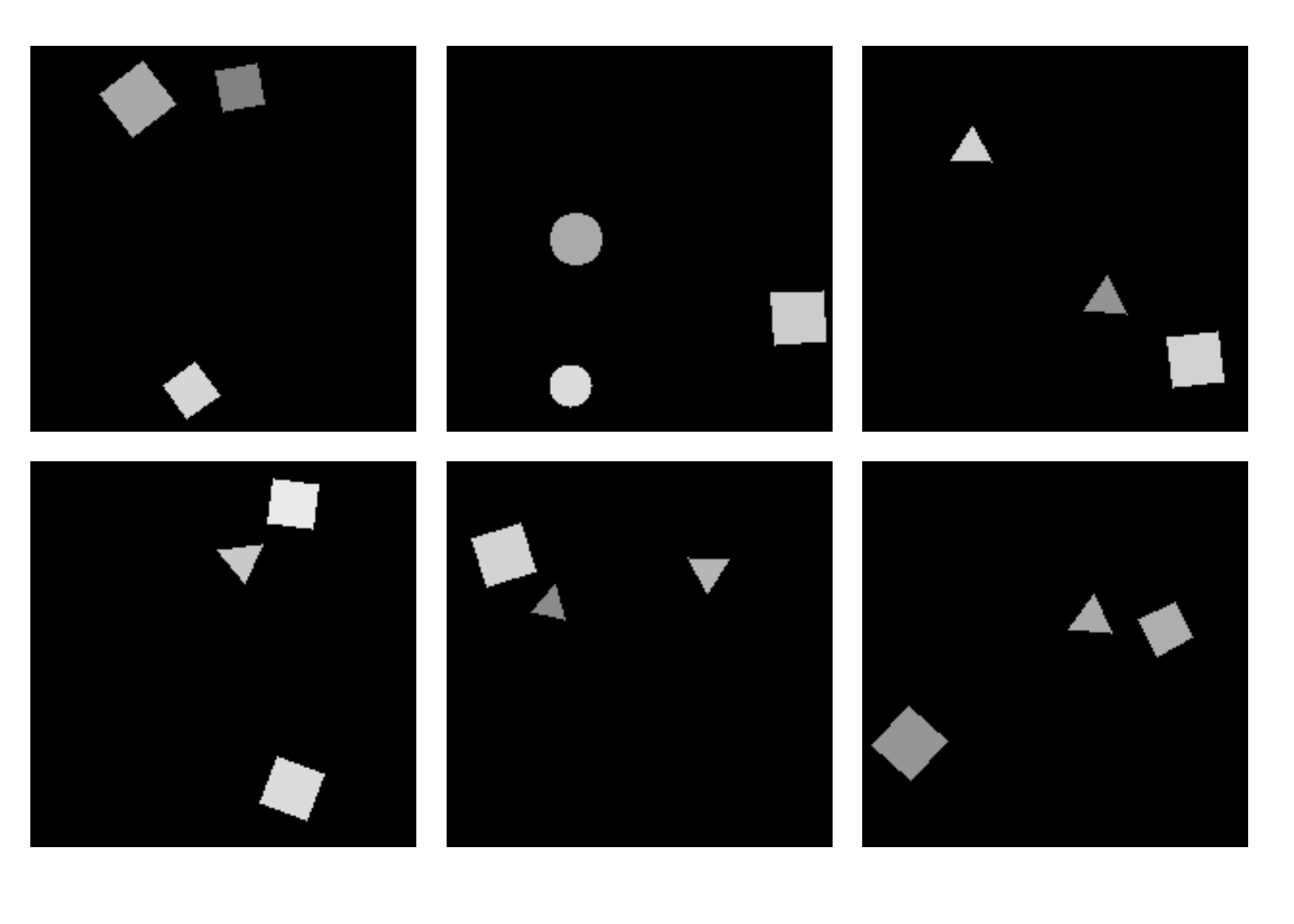}
   ~~\vspace{-10pt} 
	 \includegraphics[width=0.45\linewidth]{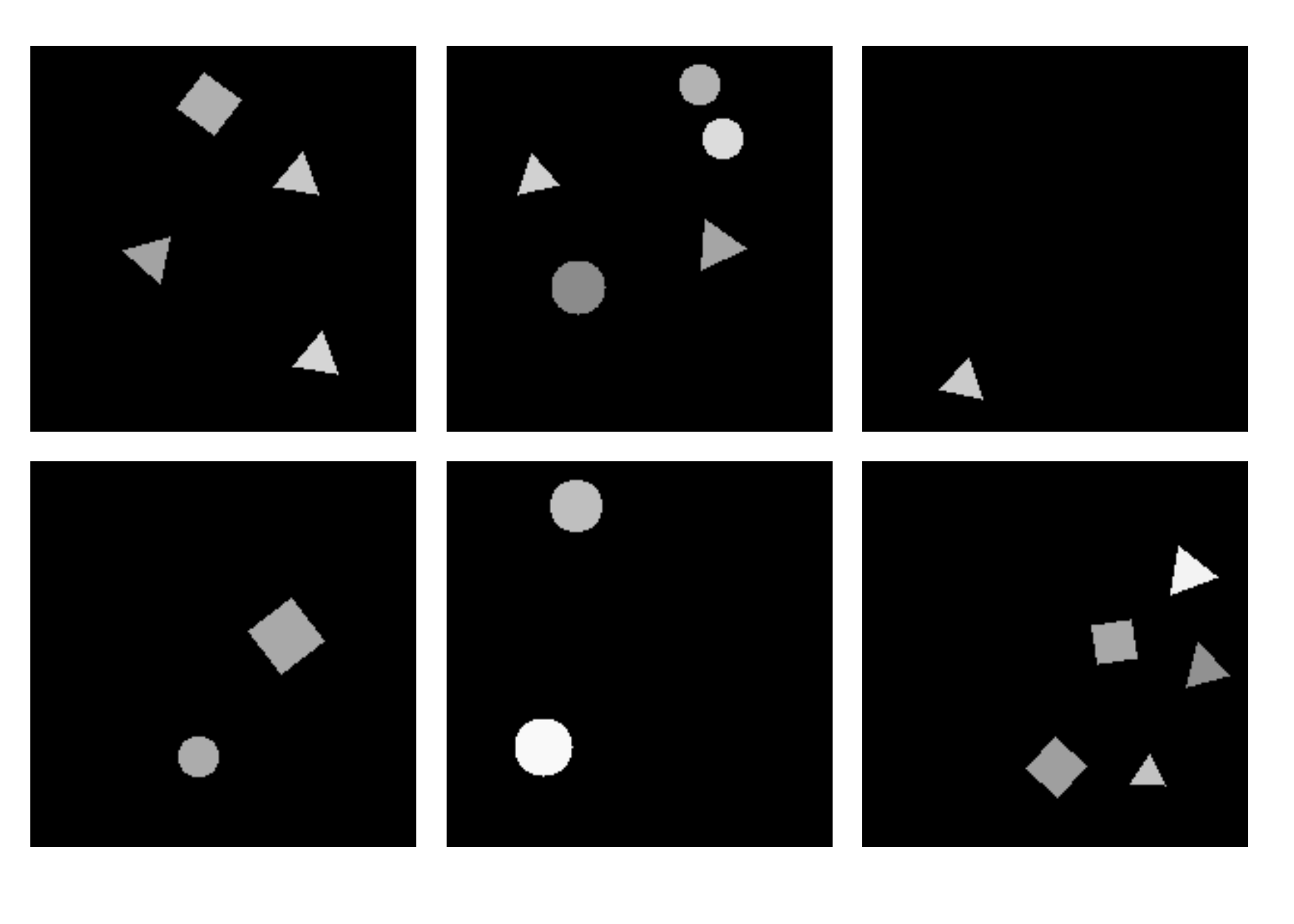}
	Setting 3\\
\end{center}
   \caption{Example of training samples for object counting task. For each of three settings: left --- 6 positive examples containing $3$ objects each; right --- 6 negative examples containing other-than-3 objects each.}
\label{fig:sample_count}
\end{figure}

Not to make the problem too trivial, a task of counting 3 objects vs. other-than-3 objects is designed. Each positive image contains 3 objects, while each negative image contains a different number (1, 2, 4 or 5) of objects. 
The experiments are conducted in three settings: in the first setting, training images only contain balls as objects. In the second setting, training objects include triangle, square and ball, but each image only contain one type of object. The third setting is similar to the second one, but each image may contain mixed types of object.
The objects could be located in various positions with different intensities and orientation. The size is in range of $[20,30]$. Examples of training images are shown in Fig.~\ref{fig:sample_count}.

Then, two deliberate testings are conducted: (1) using new objects (hexagram, F4, F2) of size $\in[20,30]$ to test shape sensitivity; (2) using the training objects of different size $\in[30,40]$ to test scale sensitivity.

\subsection{Counting types (shape uniformity/diversity)}

\begin{figure}
\begin{center}
	 \includegraphics[width=0.45\linewidth]{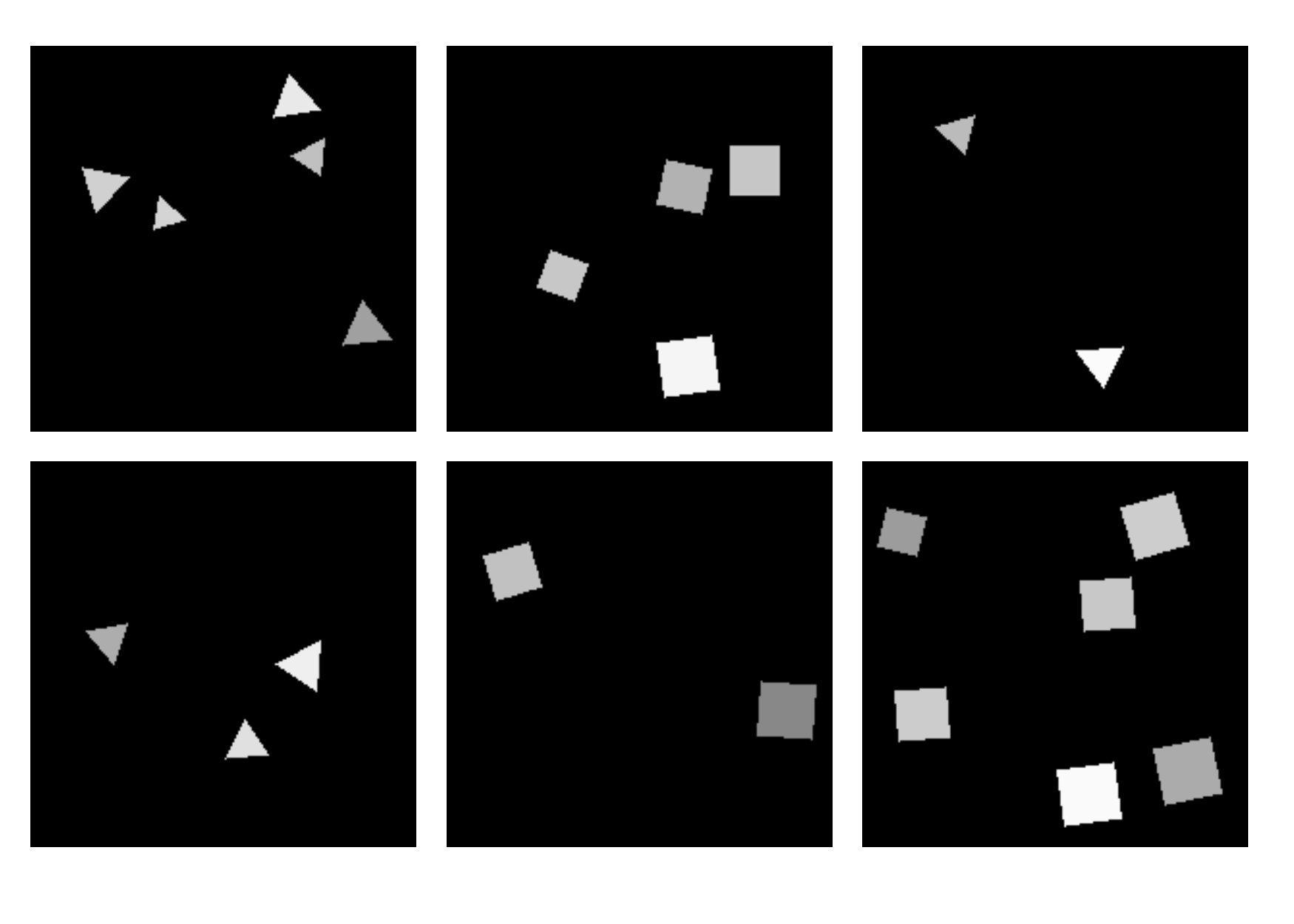}
   ~~\vspace{-10pt}
	 \includegraphics[width=0.45\linewidth]{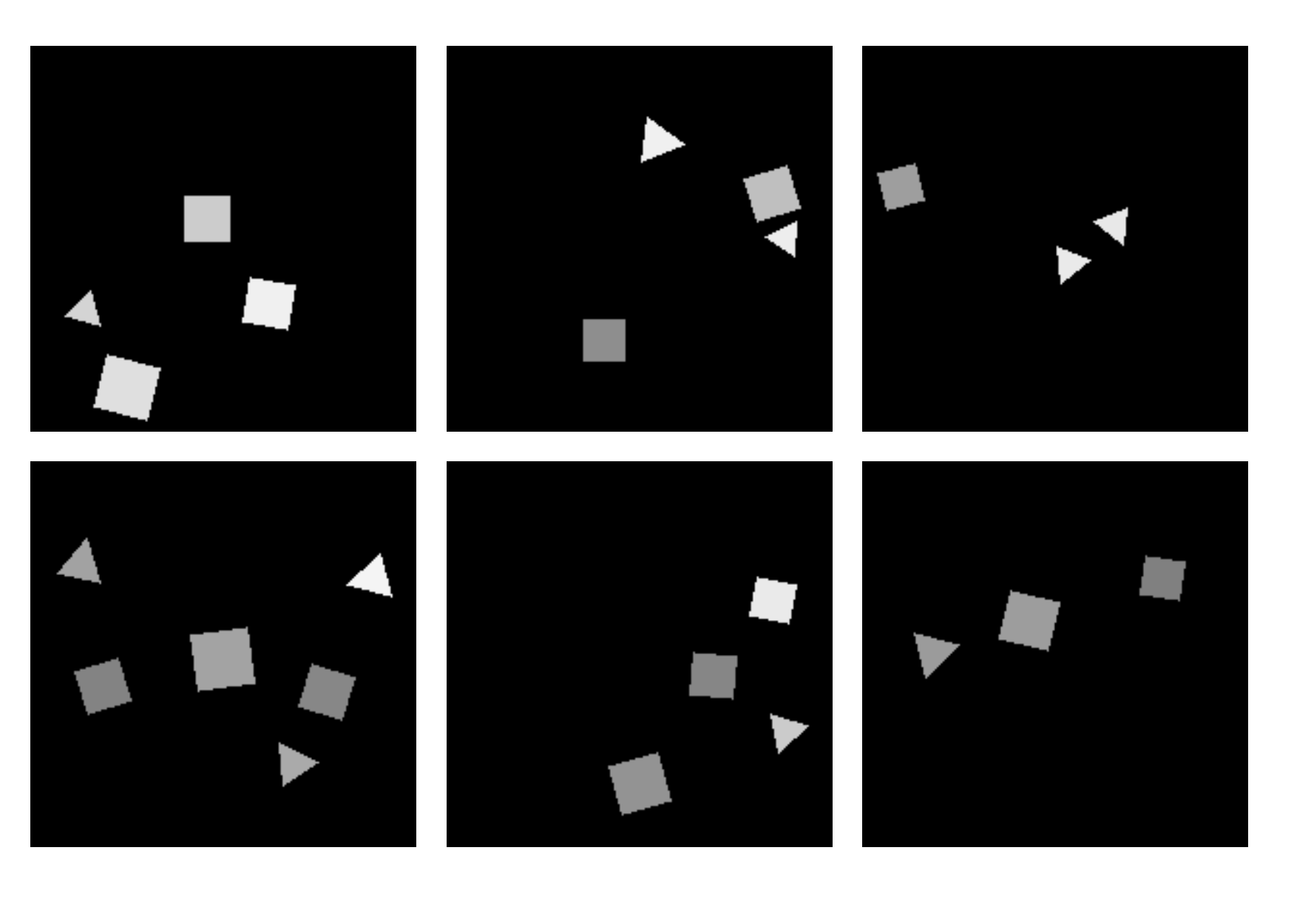}
\end{center}
   \caption{Examples for task of courting types. Left: mixture of same type of shape; right: mixture of two different types of shape.}
\label{fig:sample_mix1}
\end{figure}

In this task, we design another binary classification problem: one type of shape vs. two types of shapes. More specifically, each positive image contains multiple objects of the same type, while each negative image contains two types of objects. The number of objects in each image varies. Example are shown in Fig.~\ref{fig:sample_mix1}. This task is harder than counting objects, since it requires local shape discrimination and global reasoning at the same time.

Similar to the last section, we conduct two deliberate testings to test the generalization of the trained model. In training, we only use triangle and square to create image samples. In deliberate testing of new shapes, we use ball, hexagram and F4.

\section{Common Fate / Synchrony}
\label{sec:group}

\begin{figure}
\begin{center}
	 \includegraphics[width=0.45\linewidth]{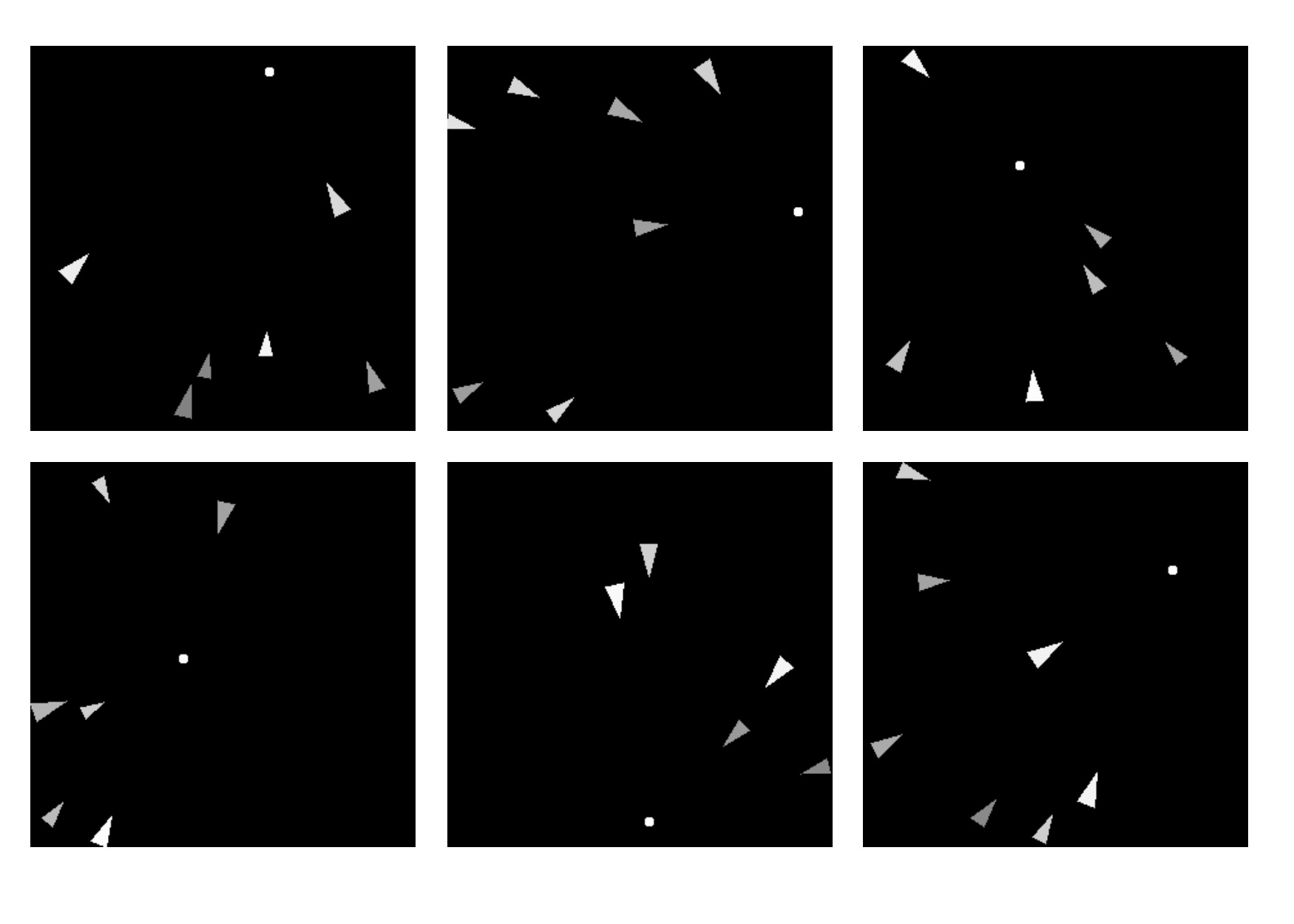}
   ~~\vspace{-10pt}
	 \includegraphics[width=0.45\linewidth]{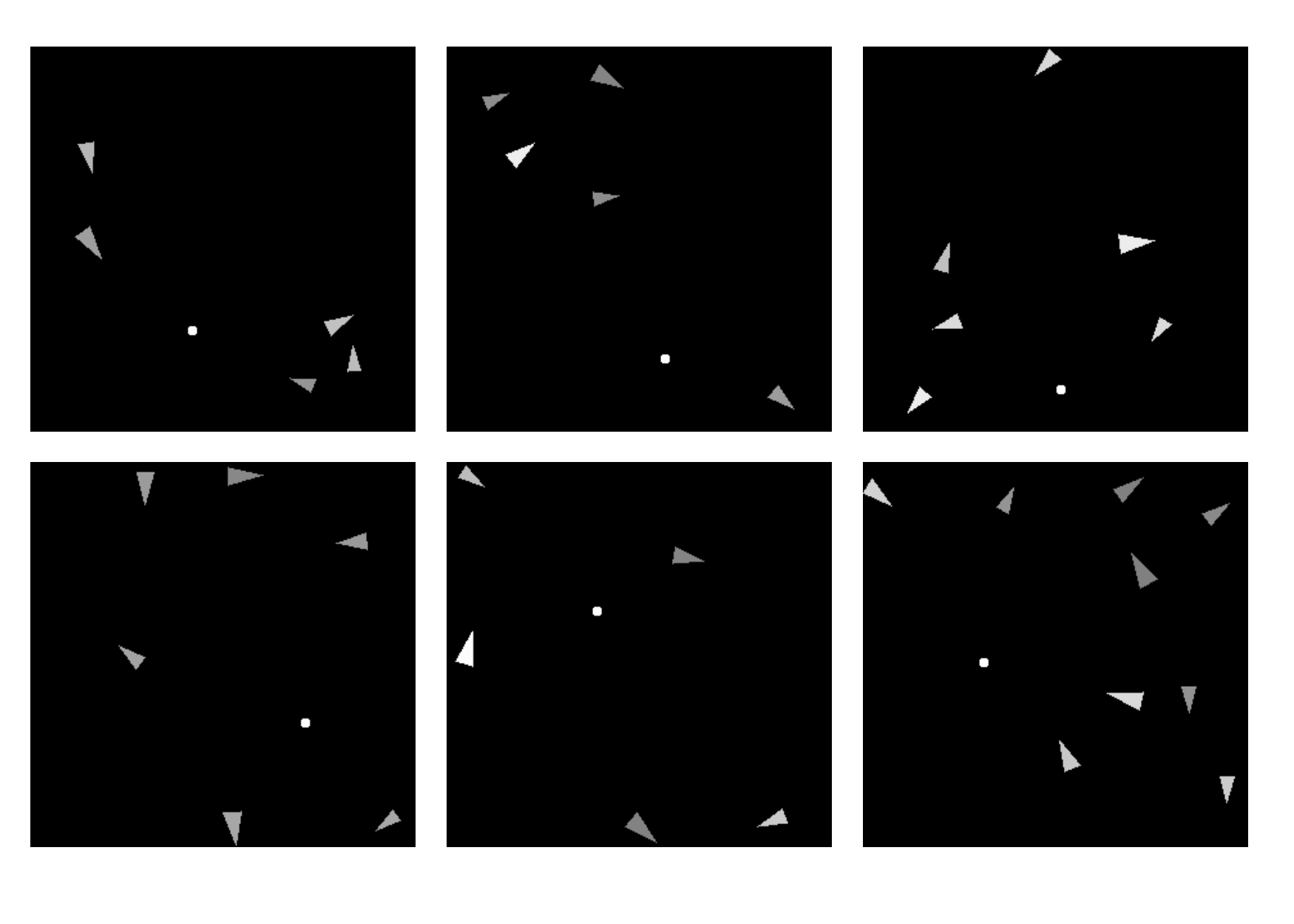}
\end{center}
   \caption{Examples for task of common fate or grouping. Left: all objects are pointing at the same target point; right: objects are not pointing at the same target.}
\label{fig:sample_target}
\end{figure}

In this section, another Gestalt-style experiment is designed to test ``grouping or conformance behavior", where multiple objects (e.g., pointy triangles) in a positive image are all behaving in a consistent manner, facing a single target (e.g., a dot). Whereas a negative image would contain objects that do not conform in the same way.
Some image examples are shown in Fig.~\ref{fig:sample_target}. 

Multiple rounds of training and testing are conducted, with the first round using randomly oriented triangles as negative images. In subsequent rounds, we test on some deliberately constructed testing samples. These could be negative samples with only a few (1 or 2) outlier objects that are not targeting the focus point; or samples with fewer or more objects; or combinations of the variations; plus size variations of the triangles.

\section{Experimental results}

Since our synthetic data sets are all gray images while the network requires 3-channel input, the images are converted to 3-channel by duplicating three times.
In all experiments, training takes 70 epochs, and model selection is based on minimum error on the validation set. Training batch size is 40 for all synthetic data sets. We report error rate (ER), accuracy (ACC), recall and precision, all in percentage.

For data augmentation, we apply randomly (1) rotation in range of 5 degrees; (2) horizontal and vertical shift in range of 2\% of image width and height; (3) flip horizontally and/or vertically.

\subsection{Results of global symmetry}

We label symmetry as class 0, and asymmetry as class 1.
In the first round, we generate $A_1$, $B_1$ and $C_1$. Each set has 8000 examples, 4000 in each class. Denote $M_1$ as the first trained model, all errors are misclassified asymmetric samples, so the recall of symmetric image is 100\%.
In the second round, model $M_2$ is trained on the enlarged datasets $A_2$ and $B_2$. Each of the data set has 16000 examples, 8000 in each class.
Similarly, in the third round training, $M_3$ is trained on $A_3$ and $B_3$, each contains 32000 examples.
We report all training and deliberate testing results in table \ref{tab:global_symmetry}.

\begin{figure}
\begin{center}
   \includegraphics[width=0.7\linewidth]{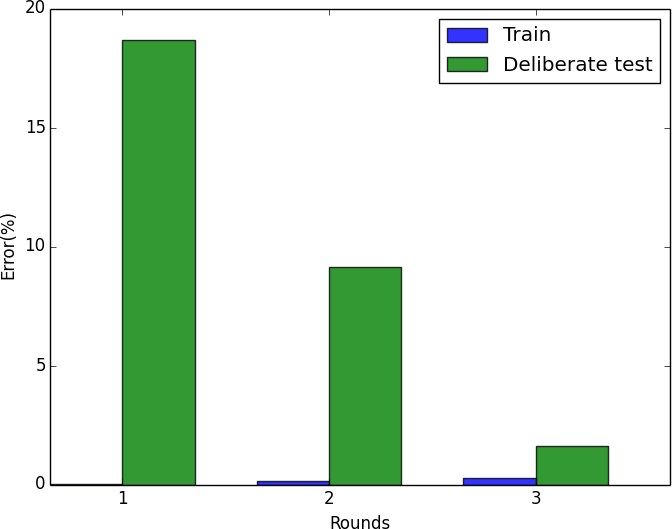}
\end{center}
   \caption{Training/testing errors of the first 3 rounds.}
\label{fig:plot_symm_error}
\end{figure}

The ER's in the three rounds of training and testing are plot in Fig.~\ref{fig:plot_symm_error}. There are two main observations from this plot. One is that as the number of training samples increases (by incorporating more deliberate samples), the testing errors become smaller. This shows the importance of large and representative training data. The other one is that the generalization of the trained model is not automatically achieved when the training set is relatively small, since the deliberate testing results are much worse than the training error. For example, the ER of $D_1(A_1)$ is about $190$ times worse than that of $A_1$. However, this is not the case in human testing, which we will report subsequently.

\begin{table}
\begin{center}
\begin{tabular}{|l|c|c|c|c|}
\hline
Model & Test Data & Accuracy (\%) & Precision (\%) & Recall (\%) \\
\hline\hline
M1 & $A_1$ & 99.98 & 99.95 & 100 \\
M1 & $C_1$ & 99.9 & 99.8 & 100 \\
M1 & $D_1(A_1)$ & 81.31 & 72.79 & 100 \\
M2 & $A_2$ & 99.84 & 99.7 & 99.99 \\
M2 & $D_2(A_2)$ & 90.85 & 84.55 & 99.98 \\
M3 & $A_3$ & 99.73 & 99.48 & 99.99 \\
M3 & $D_3(A_3)$ & 98.39 & 96.9 & 99.98 \\
M4 & $D_3(A_3)$ & 99.63 & 99.27 & 100 \\
\hline
\end{tabular}
\end{center}
\caption{Results of global symmetry in accuracy, precision and recall.}
\label{tab:global_symmetry}
\end{table}

To better understand the behavior of the trained models, we create a new set (denoted by $A_4$) with 4000 images in each class. These samples are generated by using simple shapes (triangle, square and ball) only. Some examples are shown in Fig.~\ref{fig:sample_symm5}.
1st model ER: $6.33\%$, precision 88.77\% and recall 100\%. 
2nd model ER: $0.83\%$, precision 98.38\% and recall 100\%. 
3rd model ER: $0.74\%$, precision 98.57\%, recall 99.98\%. 
We can see that the models trained in the three rounds have improved ability of symmetry recognition as more deliberate samples are included in training. It is interesting to see that almost all the errors come from the misclassified asymmetric images. This makes sense because the asymmetric patterns have much larger variance than symmetric ones.

\begin{figure}
\begin{center}
   \includegraphics[width=0.46\linewidth]{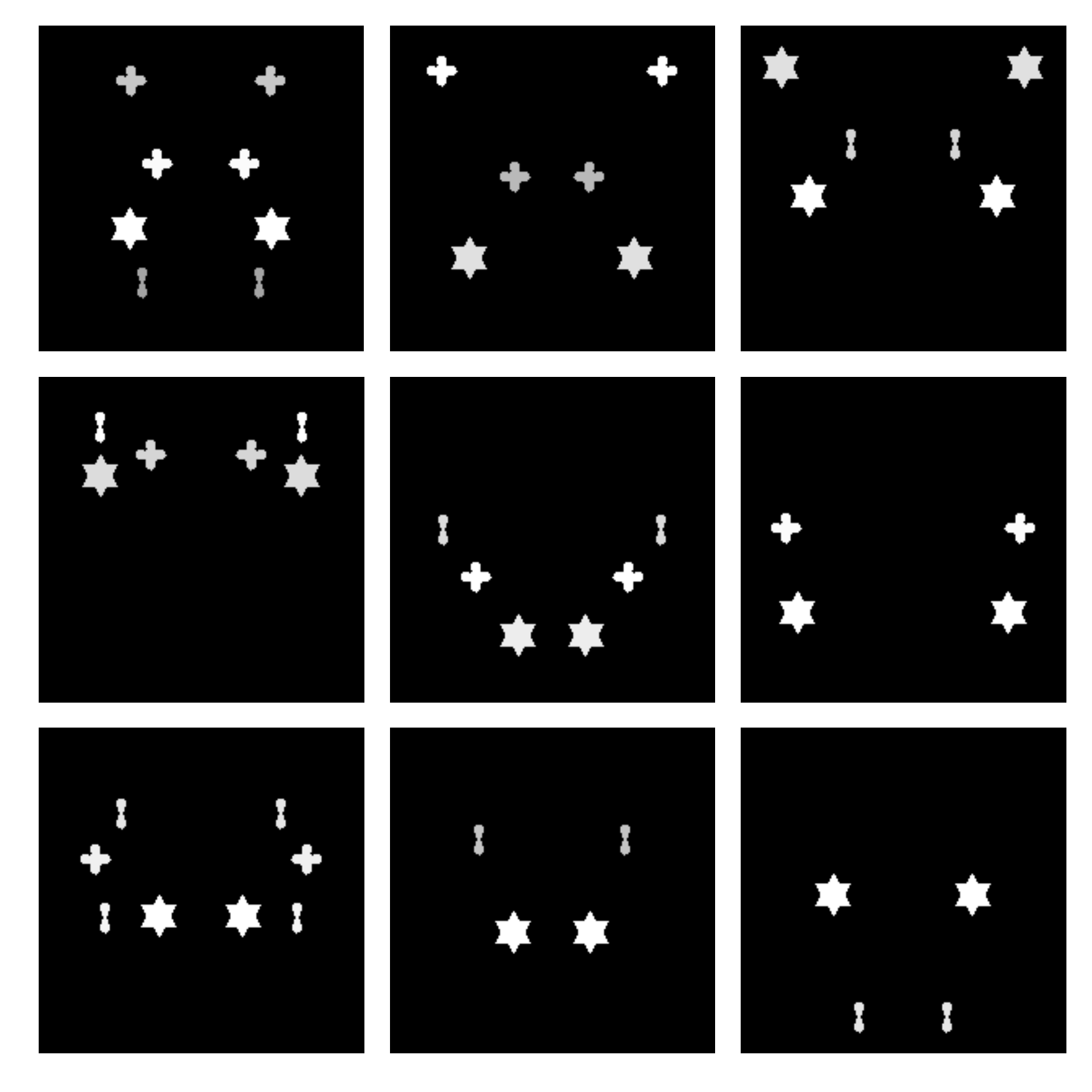}
   ~~~
	 \includegraphics[width=0.46\linewidth]{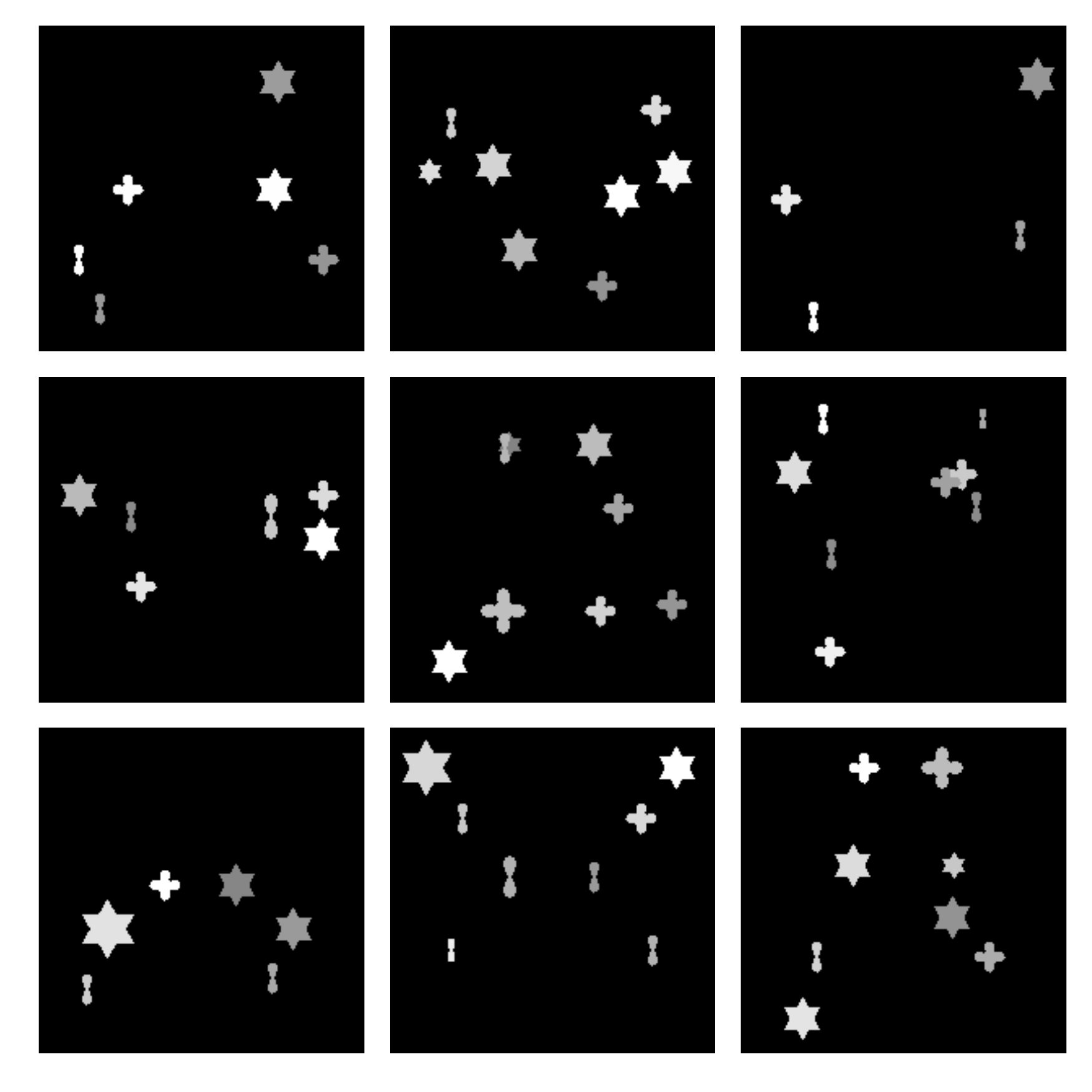}
\end{center}
   \caption{Example of final testing samples in $C_4$. Left: 9 symmetric samples; right: 9 asymmetric samples.}
\label{fig:sample_symm6}
\end{figure}

\begin{figure}
\begin{center}
   \includegraphics[width=\linewidth]{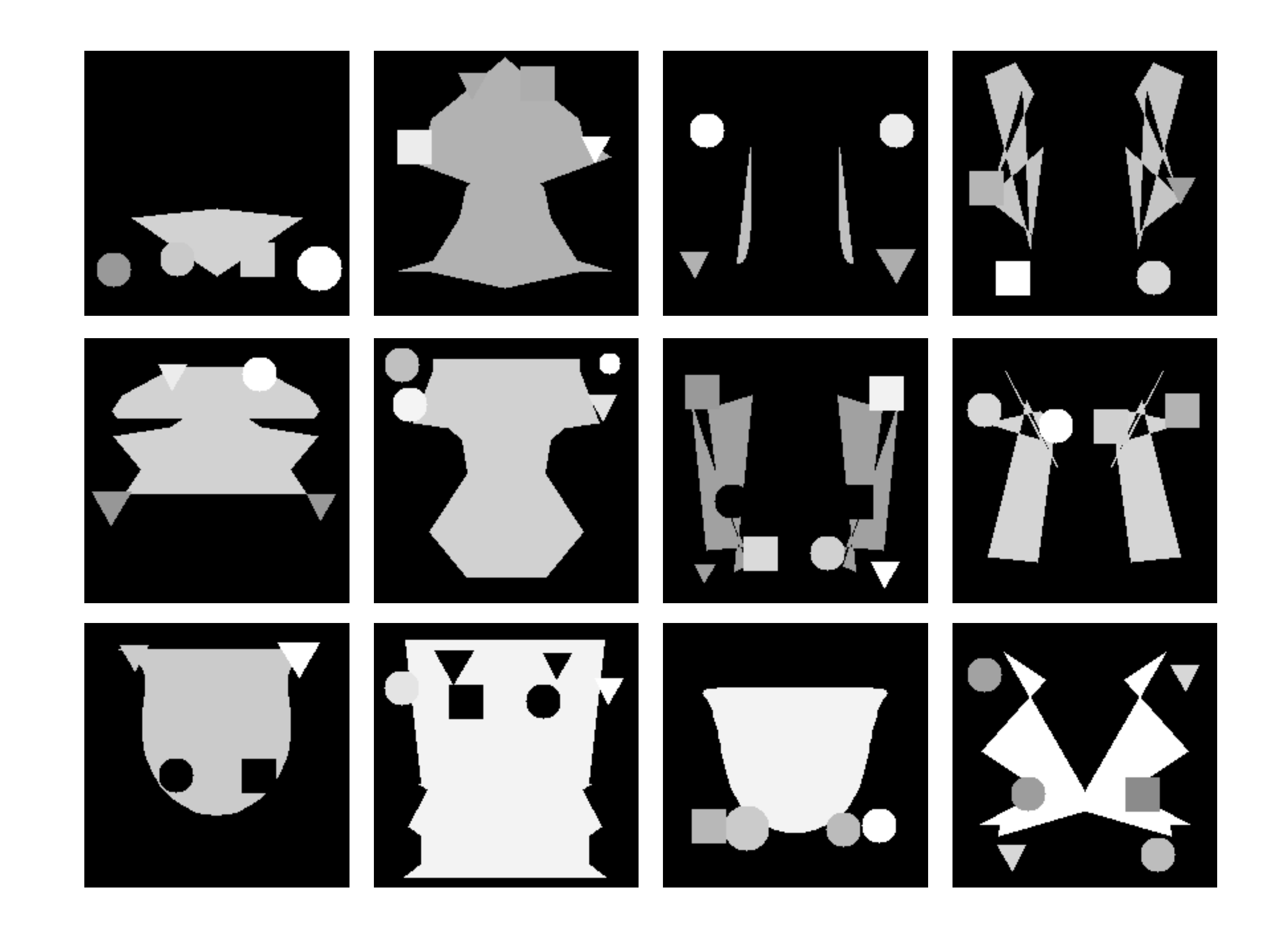}
\end{center}
   \caption{Example of error cases from 3rd deliberate test set using the 4th round model. All errors come from misclassified asymmetric samples.}
\label{fig:sample_symm_error}
\end{figure}

Finally, we conduct the 4th round of training by combining training sets $A_4$ and $A_3$, and test the final model $M_4$ on 8000 new testing samples (denoted by $C_4$). These final testing samples are created like $A_4$ by using new types of shape (hexagram, F4 and F2). Some examples are shown in Fig.~\ref{fig:sample_symm6}. The final model achieves $0\%$ ER on $C_4$.
We also test the final model on $D_3(A_3)$. The ER decreases from 1.61\% to 0.37\%.
By looking at the error cases produced by the final model (samples are shown in Fig.~\ref{fig:sample_symm_error}), we can see that the model only fail in some fine-scale details in some asymmetric images.

These final numbers, although very impressive, indicate the lack of semantic-level learning of this relatively simple concept. And the lack of sensitivity to fine-scale details is consistent with the anecdotal errors reported in the literature.

\textbf{Human test}

We developed a simple game to test human performance in the same task, which consists of maximum three rounds of training and four rounds of testing. The three sets of training samples come from $A_1$, $A_2$ and $A_3$. The four testing sets come from $C_1$, $D_1(A_1)$, $D_2(A_2)$ and $D_3(A_3)$.

To maximize consistency of the tests across different human subjects, a written instruction (see Appendix) is used and verbal communication was kept to a minimum. Each subject is firstly shown 12 pairs of positive and negative examples from two classes. A subject can request to see more examples (in increments of 3) from either class. As soon as the subject believes that he/she has learned the classification rule, we test them on 20 random examples. If succeed with no errors, a next round of testing, each with 20 random examples, will be conducted. In case the subject make mistakes in any round of testing, a new training session using corresponding training set will be given.

We tested on 15 subjects individually in two groups. The first group of 7 subjects were tested on biased training samples, in which each symmetric image has only one connected component as the foreground object. The second group of 8 subjects were tested on unbiased training samples, in which symmetric samples have more variance (can have separated symmetric objects in an image).
In the first group, 5 subjects succeeded to pass all four tests by learning from 24, 24, 30, 39, 48 examples, respectively. Interestingly, 2 subjects in the first group failed to learn the rule. These two subjects came up with different hypotheses for discriminative object properties, such as angle, roundness, connectivity, ``looks like living things?" and so on.
In the second group, all subjects succeeded to pass all tests by learning from 24, 24, 30, 33, 36, 48, 64, 300 examples, respectively.
Among 13 passing subjects, four learned from only 12 pairs of examples in the first training round; the rest required more examples from 1st or 2nd training round (made a few errors and quickly corrected them after seeing more deliberate training samples).

We found that the most tricky testing examples are a couple of the asymmetric images, in which the small and subtle asymmetry was missed by human eyes, when testing was conducted in a fast pace.


\subsection{Results of local symmetry}

We create 8000 training samples and 8000 validation samples to train the model. The local symmetric and asymmetric patterns are created in the similar way as creating $A_1$ in global symmetry task. Symmetric patterns also include small objects, like, triangle, square and ball. All foreground objects have size range $\in[30, 40]$. 
After training, 8000 statistical testing samples are created in the same manor so that they have same data distribution as the training set. 

\textbf{Deliberate test 1}: This test set is created by using new types of objects. Some examples are shown in Fig.~\ref{fig:sample_symm_pat2}.
\textbf{Deliberate test 2}: This test set is created by using the same types of objects as in training, but larger size. The new sizes range in [40, 45]. 

All results are reported in table \ref{tab:local_symmetry}. From the deliberate testing results, we can see that the learned model for local symmetry recognition is sensitive to unseen objects, but not sensitive to different object sizes.

\begin{table}
\begin{center}
\begin{tabular}{|c|c|c|c|}
\hline
Test Data & Accuracy (\%) & Precision (\%) & Recall (\%) \\
\hline\hline
Statistical test & 98.39 & 96.9 & 99.98 \\
Deliberate test 1 & 56.57 & 54.07 & 87.4 \\
Deliberate test 2 & 97.47 & 95.28 & 99.9 \\
\hline
\end{tabular}
\end{center}
\caption{Results of local symmetry in accuracy, precision and recall.}
\label{tab:local_symmetry}
\end{table}

\subsection{Results of normal vs. tampered human face}

\begin{figure}
\begin{center}
   \includegraphics[width=\linewidth]{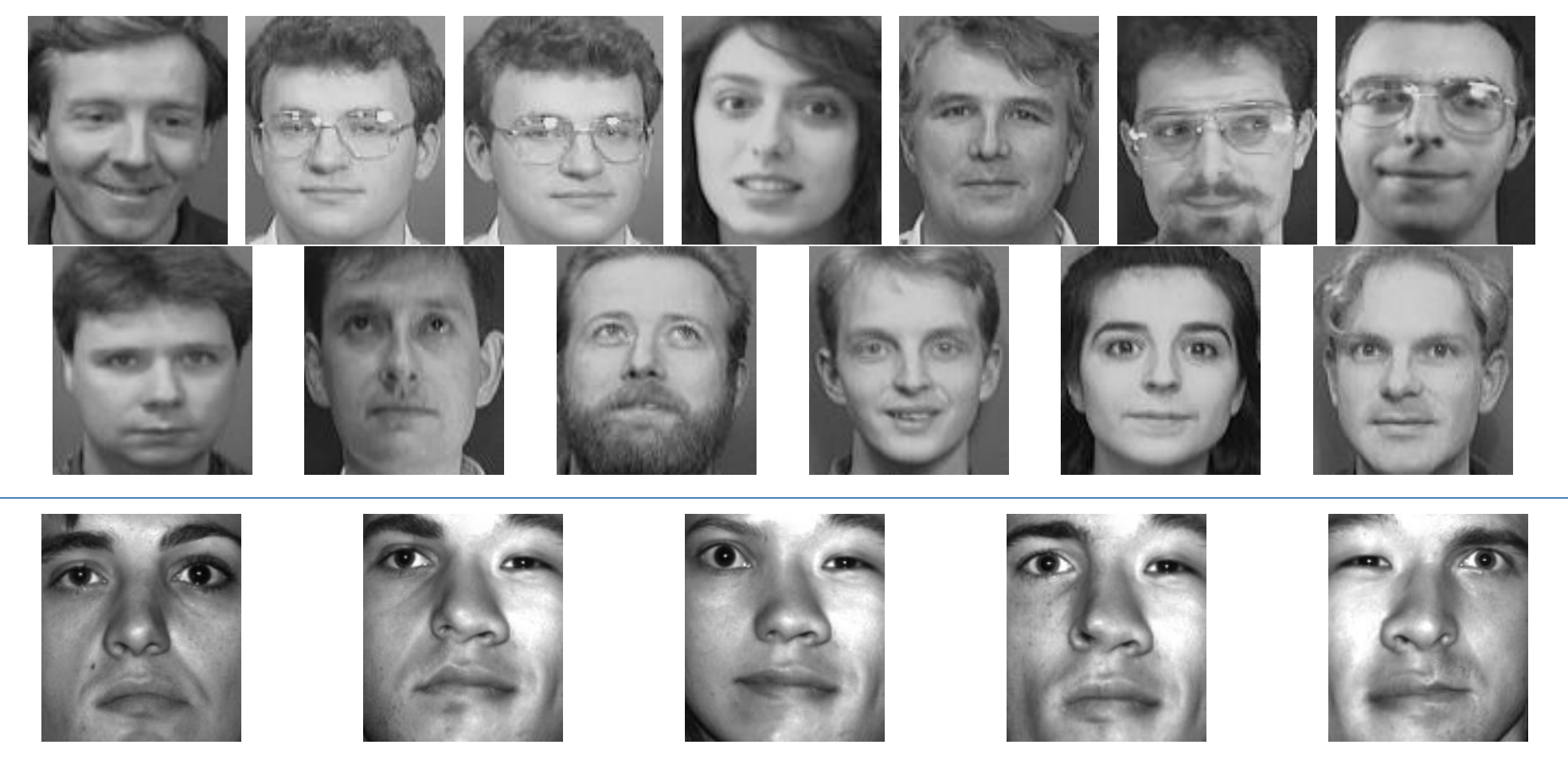}
\end{center}
   \caption{All error cases from testing set. Top samples are misclassified as tampered, while bottom samples are misclassified as real.}
\label{fig:sample_face_err}
\end{figure}

We firstly use Yale-cropped and AT\&T data sets for training and statistical testing. 
Train 1574 samples: 462 real (positive) and 1112 tampered (negative). Test 1450 samples: 338 real and 1112 tampered. Due to limited samples, we use the training set as validation set in training process. The subjects in testing set are in the same population of the training set, but the faces have varied illumination and facial expressions. For the tampered face class, two faces $I_1$ and $I_2$ are swapped to create testing samples.
Test accuracy: 98.76\%, precision 98.48\%, recall 96.15\%. Error cases are shown in Fig.~\ref{fig:sample_face_err}.

\begin{figure}
\begin{center}
   \includegraphics[width=\linewidth]{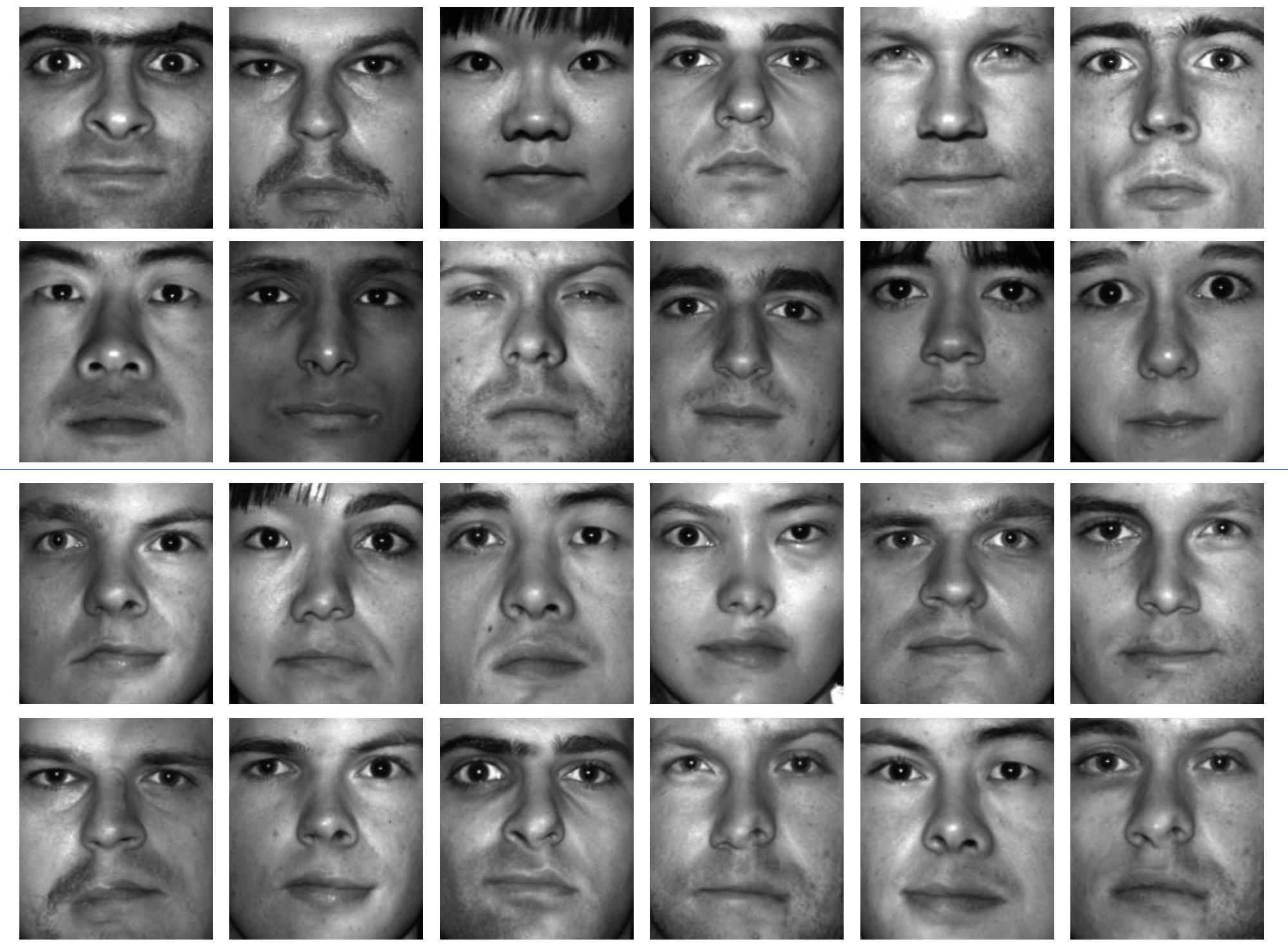}
\end{center}
   \caption{Example of error cases by pre-trained symmetry model. Top samples are misclassified as tampered, while bottom samples are misclassified as real.}
\label{fig:sample_face_err3}
\end{figure}

When we tested the best symmetry model $M_4$ from section \ref{sec:gloSym} trained on $A_4$ and $A_3$ on this face problem (only test on well aligned and cropped Yale set: 380 real faces and 1406 tampered faces), the accuracy is 81.92\%, and precision 64.47\%, recall 33.42\%. Most errors come from misclassified real faces. Random examples of error cases are shown in Fig.~\ref{fig:sample_face_err3}. It shows that the symmetry model cannot identify real faces well, which should be an easy task for human.

\textbf{Human test}

Fig.~\ref{fig:sample_face1} was shown to two children, a 12-year-old and a 14-year-old, with the question ``what is the difference between the two groups of images?" Within 20 and 5 seconds respectively, both realized that the second group of faces are all combinations from different people. We did not yet conduct the human test on a larger scale, which is part of the future work to establish more human performance benchmarks.

\subsection{Results of counting objects}

As described in section \ref{sec:count_obj}, three settings of training examples are generated (all have the same number of training samples, 2000 positive and 2000 negative) to learn three models.
Different testing sets are generated to evaluate these models. For each setting, new statistical testing samples, which have the same data distribution as the training set, are generated and tested. 
Then two deliberate testing sets are generated. 
\textbf{Deliberate testing 1}: 4000 testing samples are generated by using completely different objects (hexagram, F4, and F2).
\textbf{Deliberate testing 2}: 4000 testing samples are generated by using the same objects as training (triangle, square, ball), but the sizes are in range of $[30,40]$, which is about 50\% bigger than training shapes ($[20,30]$). 
All testing results are reported in table \ref{tab:count_object}.
For the three training settings, the ER's of deliberate testing 2 become around $1800$, $600$, $700$ times worse than that of statistical testings, respectively. 
The model trained in setting 2 outperforms that of other two settings in deliberate testing sets. The results show that training examples with more types of object are useful in this task. However, mixed types of object in one image may be distracting. With only limited number of training examples, such variance could mislead or delay the training convergence to some extent.

\begin{table}
\begin{center}
\begin{tabular}{|c|c|c|c|}
\hline
Test Data & Accuracy (\%) & Precision (\%) & Recall (\%) \\
\hline\hline
Setting 1 &  &  &  \\
Statistical test & 99.98 & 100 & 99.95 \\
Deliberate test 1 & 95.63 & 98.98 & 92.2 \\
Deliberate test 2 & 62.5 & 79.83 & 33.45 \\
\hline\hline
Setting 2 &  &  &  \\
Statistical test & 99.98 & 100 & 99.95 \\
Deliberate test 1 & 99.95 & 99.9 & 100 \\
Deliberate test 2 & 88.55 & 97.53 & 79.1 \\
\hline\hline
Setting 3 &  &  &  \\
Statistical test & 99.98 & 99.95 & 100 \\
Deliberate test 1 & 99.95 & 99.9 & 100 \\
Deliberate test 2 & 85.12 & 96.43 & 72.95 \\
\hline
\end{tabular}
\end{center}
\caption{Results of object counting in accuracy, precision and recall.}
\label{tab:count_object}
\end{table}

\textbf{Human test}

Our hypothesis is that humans can quickly discover the counting rule after given a few of the Setting 1 examples as shown in Fig.~\ref{fig:sample_count}. Once their brains are \textit{primed} with this concept, Setting 2 and 3 testing will become easy, with zero training, or at most a couple more training samples for confirmation, and the subsequent testing performance will be 100\%, with invariance to size, shape, or location etc. A quick human test of a few subjects confirmed this hypothesis. A larger scale study is left as future work.   

\subsection{Results of counting types}

In training, we use triangle and square to generate all images (4000 training and 4000 validation samples). The default size range is $[20,30]$.
\textbf{Deliberate testing 1}:
We include 3 more object types into testing, i.e. ball, hexagram and F4. In this testing set, each sample in single-type class may contain one of the three new shapes. Each sample in the two-type class includes any 2 combination of the 5 shapes.
\textbf{Deliberate testing 2}:
We also test on new samples with larger objects. Two sets are created with size range $[30,40]$ and $[40,50]$, respectively. 

\textbf{Additional test 1}:
Similar to global symmetry task, we can add new samples to train new model, which is expected to have less over-fitting and better generalization.
We add F4 into training shapes to train a new model and test on the same testing set as in deliberate testing 1. 
\textbf{Additional test 2}:
Following deliberate testing 2, we train a new model using samples with shapes of small and large sizes, specifically sizes range in $[20,25] \cup [40,45]$, and test on size [30,35] to see whether there is improvement of generalization. 

Results are reported in table \ref{tab:count_type}, which show that the learned model is sensitive to new object shapes and scales. Even after additional training with new training samples, the models still do not get the logic of type counting.

\begin{table}
\begin{center}
\begin{tabular}{|c|c|c|c|}
\hline
Test Data & Accuracy (\%) & Precision (\%) & Recall (\%) \\
\hline\hline
Statistical test & 99.98 & 99.95 & 100 \\
Deliberate test 1 & 73.67 & 76.68 & 68.05 \\
Deliberate test 2 [30,40] & 86.6 & 97.66 & 75.0 \\
Deliberate test 2 [40,50] & 56.7 & 78.39 & 18.5 \\
Additional test 1 & 83.1 & 79.53 & 89.15 \\
Additional test 2 & 72.75 & 69.36 & 81.5 \\
\hline
\end{tabular}
\end{center}
\caption{Results of type counting in accuracy, precision and recall.}
\label{tab:count_type}
\end{table}

\subsection{Results of common fate}

\begin{figure}
\begin{center}
	 \includegraphics[width=0.45\linewidth]{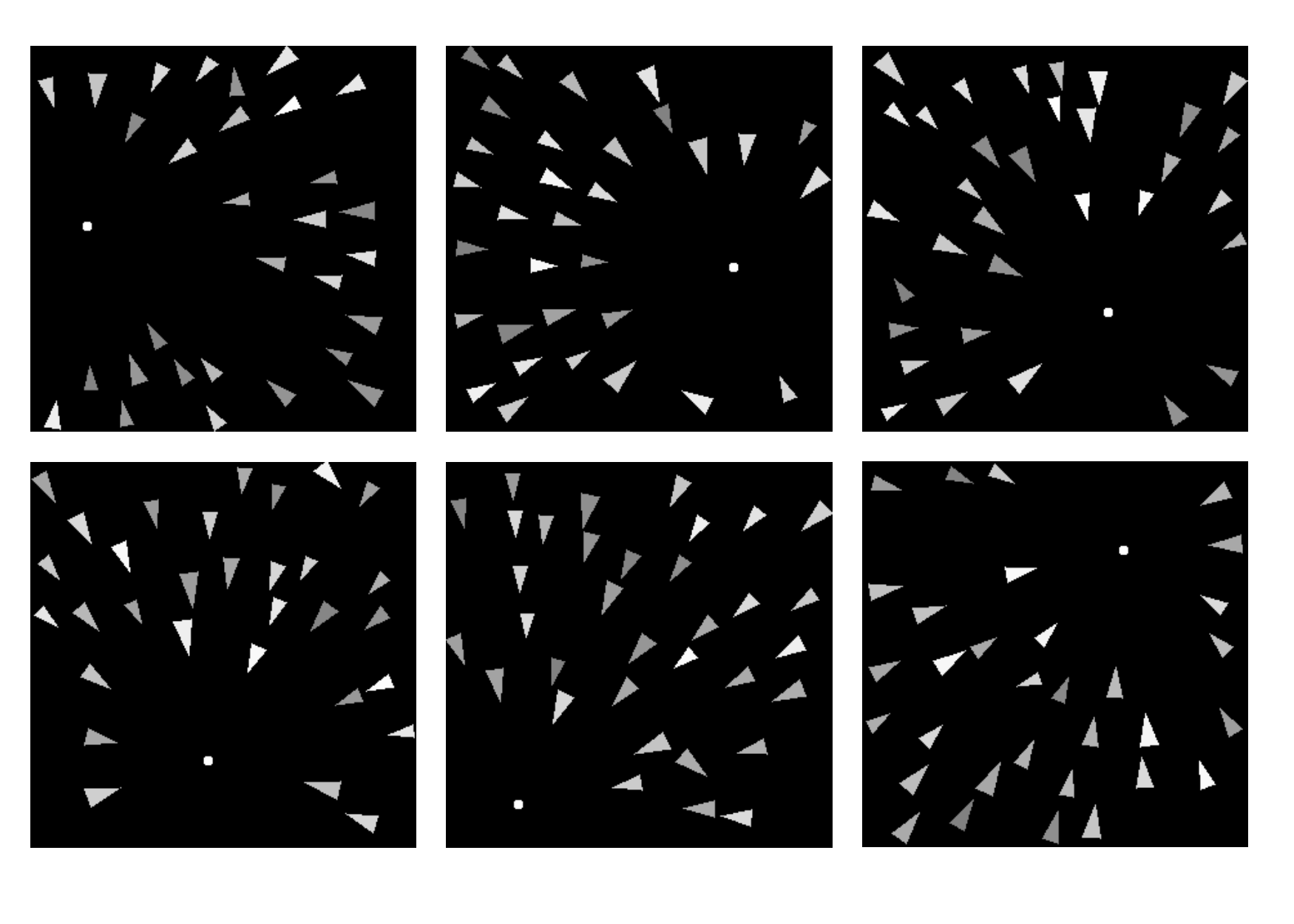}
   ~~\vspace{-10pt}
	 \includegraphics[width=0.45\linewidth]{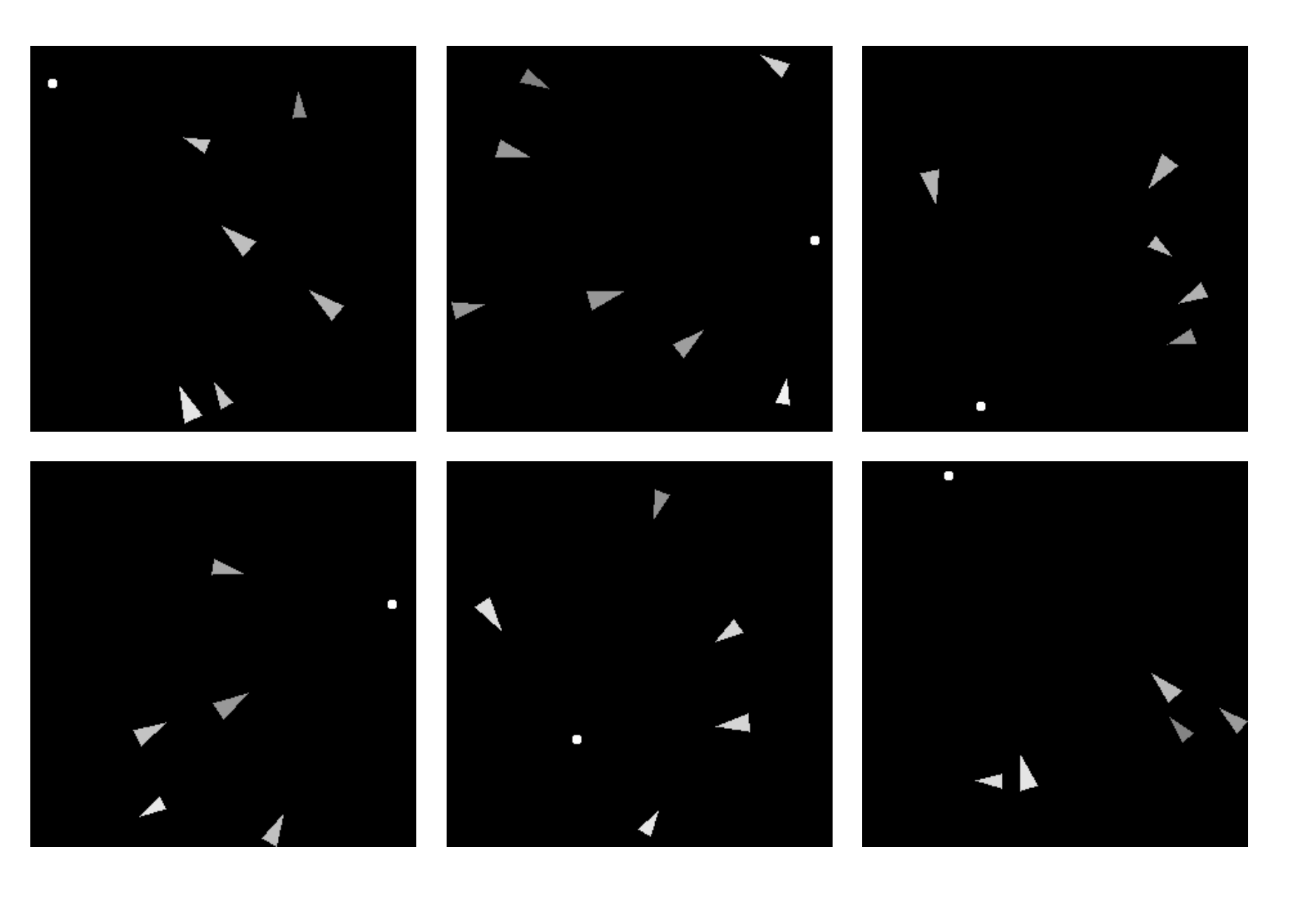}
\end{center}
   \caption{Examples of additional negative images in the 3rd and 4th rounds. Left: negative images with more objects and 1 outlier in 3rd round; right: negative images with less objects and 1 outlier in 4th round.}
\label{fig:more_sample_target}
\end{figure}

We trained 5 models in sequential rounds, each time adding some new variations of image samples into the training set. The first dataset contains the samples (4000 training, 4000 validation) similar to Fig.~\ref{fig:sample_target}, but each image has $n\in[10,17]$ objects and one target point. In negative samples, all objects are facing random directions. In the second set, 4000 new samples are added in both training and validation, where all objects except one or two outliers in the negative images are facing the target (outliers are facing at least 60 degrees away from the target). In the third set, new samples are added again, where all images contain $n\in[30,34]$ objects, and negative images each have only one outlier. In the fourth set, new samples similar to the last round are added, but each with fewer objects ($n\in[5,7]$). Some negative image examples are shown in Fig.~\ref{fig:more_sample_target}. In the fifth set, the number of training/validation samples are doubled. In each training round, the best model was picked based on accuracy on the validation set.

In this task, we observed the same behavior as in the other tasks, where deliberate tests yielded clear accuracy degradation, and training with deliberate samples would bring the accuracy back up. Below we report a different set of accuracy numbers, all based on the \emph{same} hold-out test set, which is different from all the training sets. 

\begin{figure}
	\begin{center}
		\includegraphics[width=0.9\linewidth]{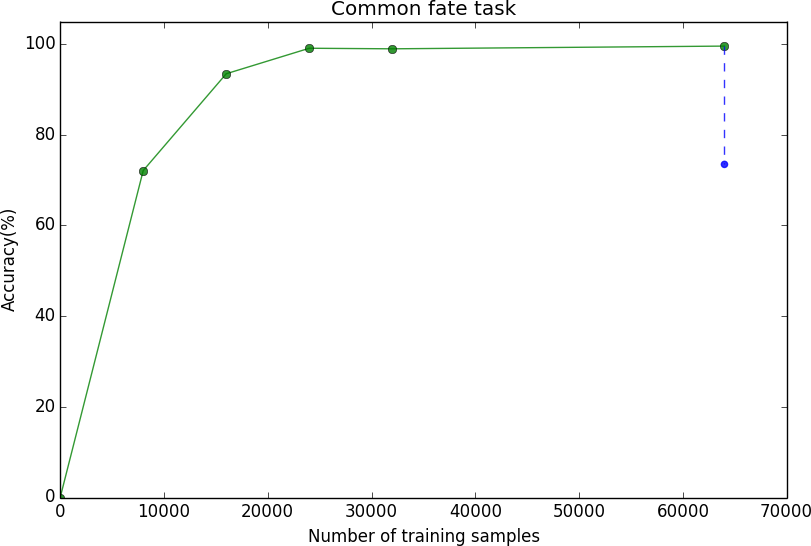}
	\end{center}
	\caption{Learning curve (green) of Inception-V3 in common fate task based on the hold-out testing set. The blue point indicates the performance on the altered test set with larger triangles.}
	\label{fig:learn_curve_target}
\end{figure}

\begin{figure}
	\begin{center}
		\includegraphics[width=0.7\linewidth]{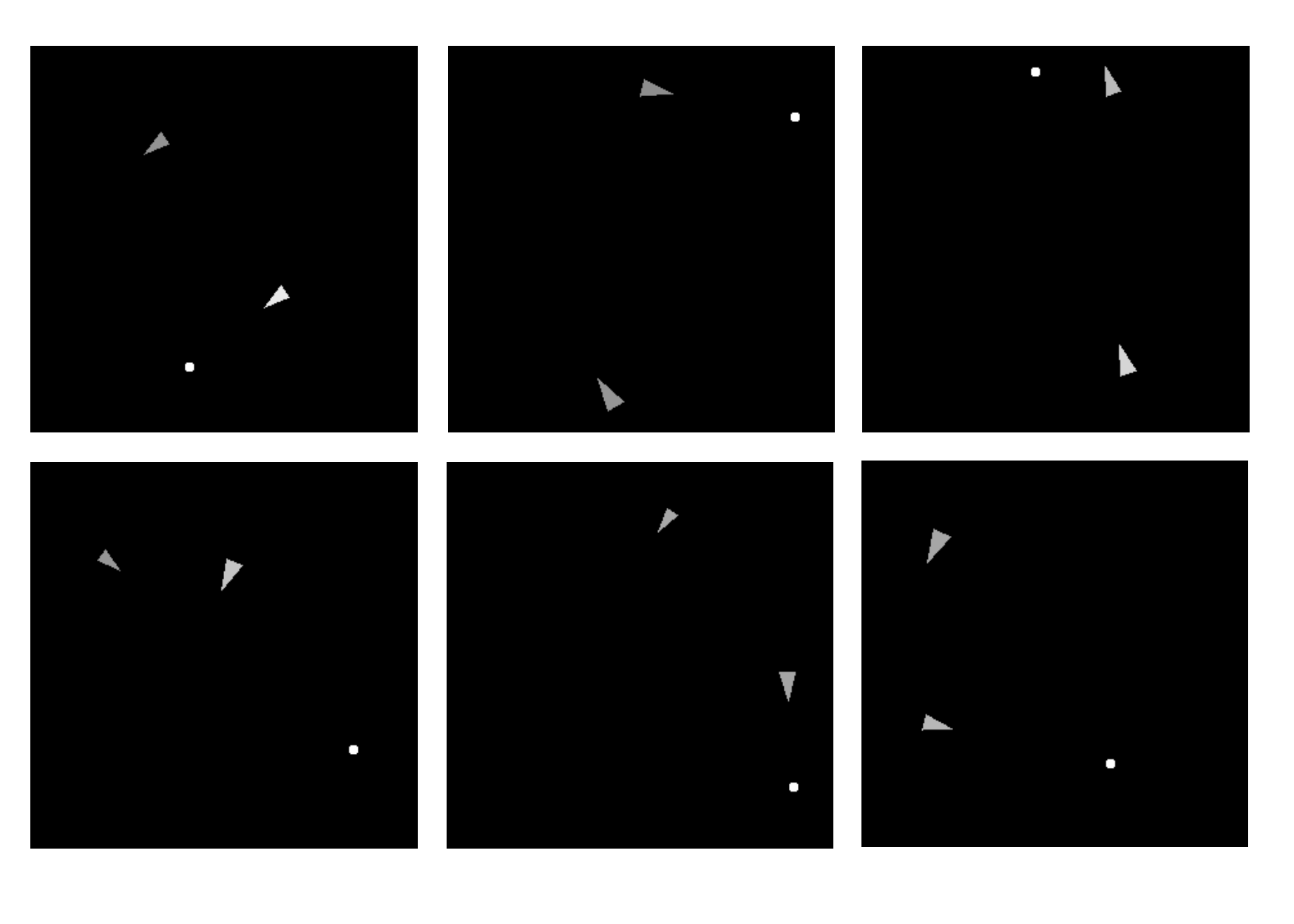}
		~~\vspace{-12pt}
	\end{center}
	\caption{Error cases on the hold-out test set by the 5th model.}
	\label{fig:errors_target}
\end{figure}

The hold-out testing images are more sparsely populated than all the training images, with each image containing only two objects and each negative sample has one outlier. This hold-out set was used to evaluate the performance of all trained models, in order to comparably observe the learning progression. The learning curve (green) is plotted in Fig.~\ref{fig:learn_curve_target}.
The learning curve shows a promising accuracy ($99.58\%$) for the final model trained and validated with 64000 images. However, the model still makes errors. Some of the error cases are shown in Fig.~\ref{fig:errors_target}.


Finally, we altered the hold-out test set, to make all triangles bigger (by doubling their size). The testing accuracy dropped to around 74\% (the blue point in Fig.~\ref{fig:learn_curve_target}). Interestingly, \emph{all} errors are false negatives.

\textbf{Human test}

Seven human subjects participated in the classification task (using the first set). All subjects used less than 20 training samples to reach the Aha-moment of grasping the semantic concept. 

One of the subjects was tested for unsupervised learning as a first step. Four training images, two positive and two negative, are shown to the subject with the question ``can you separate these into two groups?" The human subject succeeded in both the clustering task as well as the subsequent classification task (``can you put these images into the two groups you just identified?”), with 100\% accuracy.

\section{Discussions}

\subsection{The intelligence gap}
It seems that in at least two aspects machine learning has not closed the ``intelligence gap": one is that it needs many more training samples than humans do; and the other is that the ``aha moment" seems to be still uniquely human, and machines have not show any sign of mastering this trick. 

\begin{figure}
	\begin{center}
		\includegraphics[width=0.9\linewidth]{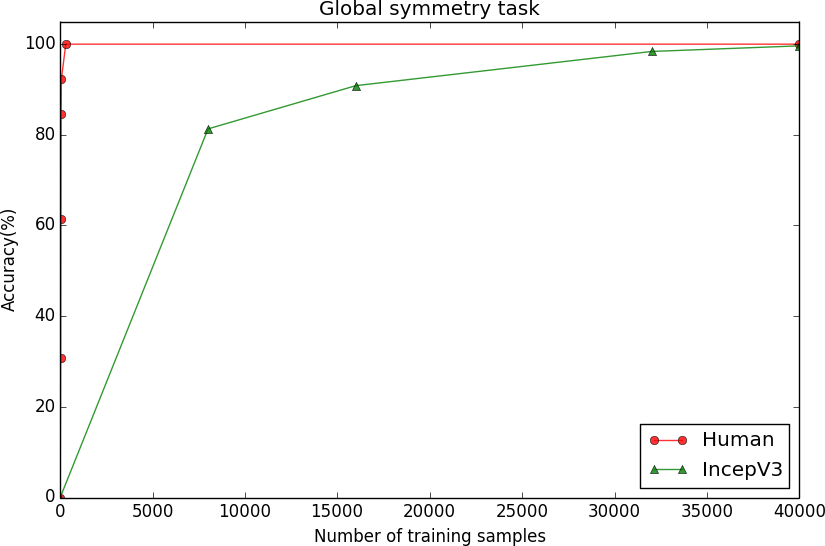}
		~~\vspace{-5pt}
	\end{center}
	\caption{Learning curve of human vs Inception-V3 in global symmetry task.}
	\label{fig:learn_curve}
\end{figure}

The comparison of learning curves by human and deep learning in symmetry task is shown in Fig.~\ref{fig:learn_curve}. The human accuracy is calculated as the percentage of head count who reach the ``aha moment", at sample size 24, 36, 48, 64 and 300. We can see that human can learn the rule from examples quickly and come to the ``aha moment" after seeing very few examples. On the contrary, DCNN can only statistically approximate the bilateral symmetry after a large number (40,000) of training examples, and still cannot reach $100\%$ accuracy on unseen data.

Artificial Neural Networks have indeed shown more sophisticated ``intelligence" than counting. One example was demonstrated by Hoshen and Peleg \cite{hoshen2016visual}: they were able to get a simple Neural Network to learn the mathematical operation of addition from pictures of numbers as inputs. Considering that human children typically learn counting first before they are taught how to add numbers, this achievement is quite impressive. Nevertheless, DCNN's rather rudimentary visual counting ability serves as a reminder that there is still some way to go to achieve a basic but semantic level of artificial visual intelligence.

\subsection{Study implications}
Imagine that you have access to anonymized brain scans from 10,000 patients. 2000 of them eventually died of a certain disease, and the rest lived long and were free of that disease. Now you want to find out whether the brain scans could have predicted the disease. You could try an ``end-to-end" learning using DCNN by feeding the brain scans as training examples. Will it work? Or will you need to understand more about the disease and its (potentially very complex) manifestations in the brain scan?

Let's assume that the disease manifestation in the brain is actually based on different configurations of some ``hot spots" (i.e., high intensity areas), with an asymmetric configuration indicating disease, while symmetry predicts no-disease. Or, the number of hot spots within a certain region of the brain is indicative of the disease: having, say, three or less hot spots means healthy, and more means disease, regardless of shape or size of each hot spot. Or, it is the diversity/heterogeneity of those hot spots that mattered. Or, it is a combination of these rules for different parts of the brain that are working together to determine the patient's fate. But you do not know any of these ``rules" \textit{a priori}. 

Based on the experiments in this paper, you have reason to stop and think, and to doubt that a direct application of DCNN will \textit{always} solve the problem. And you know roughly how many more examples you may need if you do suspect a particular rule is at work. Or, at least you know how to design synthetic experiments to shed some lights on this problem, e.g., to look for a performance limit, in case you have failed a first round of ``end-to-end" learning.

\subsection{Is visual intelligence learned or hard-wired?}
One might argue that symmetry is not a fair test case, because humans may be hard-wired to be sensitive to it. Indeed, symmetry is very common in our world, both in natural anatomy and in man-made objects. And being sensitive to symmetry probably carries a strong evolutionary advantage. But then again, one can also argue that most, if not all, of human visual intelligence may be somewhat hard-wired. If this is true, then does it mean that an ``end-to-end" learning-from-examples paradigm may not be sufficient to achieve full visual intelligence? And that proper understanding, followed by direct hard-coding (i.e., semantic modeling), of such wirings may be at least an indispensable stepping-stone?

\subsection{Why synthetic testing?}
Another question could be, why do we need synthetic examples, while we have already large annotated data sets like ImageNet? We believe popular learning targets such as cats, dogs, human faces, or more generally, animals on earth, are not the best targets for testing a learning machine rigorously. Animal species today are very sparsely distributed on the evolutionary continuum due to multiple prehistorical mass extinctions---there is no other animals with zebra stripes, or any other animal with a long nose like an elephant. When a learning machine sees a striped table cloth and calls it a ``zebra", or calls a child wearing a long-nosed mask an ``elephant", is it wrong?---After all, the table cloth could be a cutout from a massive zebra picture, and ``elephant" would be the exact right name to call a child during her Halloween adventure. Therefore, when a CNN is fooled by some noise-like adversarial overlays, could it be that it is seeing a ``zebra stripes"- or ``elephant-nose"-type of unique feature for that object? What really sets humans apart is that humans see the \textit{full picture}: the table and the child together with all the other distractions. In order to see the full picture, then, CNNs have to learn \textit{all} the objects and concepts at the same time. This is not yet possible today.

In this paper, we step back and investigate at a much smaller scale --- we focus on simple and well-defined concepts, and clean and synthetic examples to test and compare the ``intelligence" of algorithms and humans, in a quantitative way.  


\subsection{Error analysis}
For symmetry tasks, DCNN made more mistakes in classifying asymmetric patterns as symmetric ones. This behavior is somewhat like that of a human. The reason may lie, on the one hand, in the fact that symmetry is a precisely defined concept in which even a one-pixel-difference would make it asymmetric; and on the other hand, in the limited ability for fine-scale discrimination, of humans as well as of machines. Humans often make careless mistakes or suffer from sensory defects (e.g.,short-sightedness); while today's DCNNs are not sensitive to fine-grained variations in the images, unless re-designed specifically for a particular task \cite{lin2015bilinear}.

In the local symmetry study and facial study, the DCNN models learned very well on seen shapes/distributions. However, it demonstrated limited generalization in terms of new objects (second row in Table \ref{tab:local_symmetry}), even though most of the errors should be very easy for human to avoid (e.g. error cases in Fig.~\ref{fig:sample_face_err}). 

The counting experiments show that DCNN is too sensitive to object scale in the object-counting and common fate task; and failed to generalize in scale and object type in the object-type-counting task.
With limited amount of training samples, the DCNN model cannot learn the counting rules precisely. It tends to learn a small kernel space that could just cover the training sample distribution. For example, in the last scale testing experiment, the learned model can learn to fit the small and large scales, yet when tested at medium scale, the error rate spikes up. Similarly, objects with a new unseen shape seem to be able to easily confuse the learned models. In the common fate task, we see clearly that scale-invariance is not achievable without exhaustive enumeration or explicit modeling/coding. 

Humans have the prior knowledge (or tendency) that each connected component is counted once, regardless of its shape or size variations. Our trained DCNN model does not have this explicit knowledge or tendency. So, it tends to, interestingly, count twice or more for objects larger than what it saw in the training set. But can we really judge that this behavior is wrong? or merely ``uninformed"? I.e., uninformed of our world in which scale-invariance is a common law due to both the natural growth phenomenon and our perspective visual system. Imagine an alien world where things grow in density instead of size, and a sensing system based on parallel waveforms or touch only, then scale-invariance would be a very alien concept. This brings the question: how much of our visual intelligence is specific to this world that we live in, and to this bodily construct that we possess? and not \textit{universal} at all?

In the end, a machine may have to live among us to learn the whole experience, before acquiring Gestalt-style-intelligence in visual perception.

\subsection{Sample size and general intelligence}

Our experiments show that some ``intelligence gaps" can be \textit{asymptotically} closed with sufficiently large training data, albeit at very different convergence rate. One can hope that refinement of the network architecture, trained with ever larger data sets, will bring about an eventual breakthrough. However, others may believe that ``general intelligence", the kind of intelligence that understands relationships and complex rules, and that sees the full picture, can be reached only by tackling the Gestalt-type ``aha moment" problem (plus the Winograd Schema Challenge~\cite{levesque2011winograd}) first. Our object-counting experiments support the latter view. For one thing, a DCNN can learn that an object of many different sizes is the same thing as long as examples of each size is present in the training set. But it could never learn or infer by itself the rule that ``size does not matter!"---this rule has to be hard-coded.

Even if a data-driven deep learning algorithm can learn complex rules in a given domain, it may still have a difficult time combining rules or prior knowledge from multiple different perspectives (e.g., ``size does not matter", ``rotation does not matter", but ``shape matters"). With an increasing number of potentially hidden rules in effect, the required sample size for ``end-to-end" training will grow exponentially. 

\subsection{Future work: Aha Challenge, a call for participation}


The datasets used in this work and scripts to generate the synthetic images are available online at https://github.com/zhennany/synthetic.

As future work, we would like to propose an ``Aha Challenge" to the research community. The idea is to push together focused and quantitative research on \textit{algorithmic} vs. \textit{natural} visual intelligence, toward Gestalt-style pattern recognition based on a relatively small number of training images.

The goal of this challenge is two-fold: to achieve a deeper, and quantitative, understanding of the nature of human visual intelligence; and to improve higher- or semantic-level ``intelligence"  in visual learning algorithms. 

Participation can be in three forms: submission of new algorithms; submission of human study results; and proposal for new types of tests.

An algorithm submission should be tested on multiple visual learning tasks, each designed with deliberate adversarial testing that a human can easily pass (once an ``aha moment" has been reached). 

A human subject study should be conducted as comparable as possible to the algorithm training process, with minimal extra information or hints provided, besides showing the training images to the human subjects.


This challenge is analogous to the "Winograd Schema Challenge"~\cite{levesque2011winograd} in the language understanding domain. As demonstrated by the Winograd Challenge, if the machine does not have all the knowledge of this human world, it cannot correctly infer all the meanings of, or relationships among, words or visual objects.

We invite submissions of either new types of tests, new algorithms, or human study results, to establish ``intelligence" baselines, and to build machines that one day will be able to exclaim ``Aha! I see!". 


\section*{APPENDIX: Written instruction for the human test}

Below is the written instruction used in the human test for the symmetry study. 

\begin{quotation}
\begin{em}
You are participating in a game of ``visual classification''. The goal for you is to learn to classify visual patterns into two classes, after seeing examples of both classes.

Here is how the game will proceed:

1.  You will be shown examples of visual patterns from two classes, class 0 on the left of the screen and class 1 on the right;

2.  As soon as you believe that you have learned the classification rule, you can stop the training;

3.  You will be shown 20 random test examples. Please label each as either class 0 or class 1;

4.  If you succeed with no errors, 3 additional rounds of testing, each with 20 patterns, will be conducted to confirm your learning;

5.  In case you make mistakes in any round of testing, a new training session will be given to strengthen your learning. Again you can stop the training once you believe you have learned the classifier;

6.  The game stops after a maximum of 4 rounds.

Please keep this confidential so that we can invite more participants.
\end{em}
\end{quotation}



{\small
\bibliographystyle{elsarticle-num}
\bibliography{DL_ability}
}

\end{document}